\documentclass[10pt,journal,compsoc]{IEEEtran}

\ifCLASSOPTIONcompsoc
  \usepackage[nocompress]{cite}
\else
  \usepackage{cite}
\fi
\ifCLASSINFOpdf
\else
\fi
\usepackage{amsmath,amsfonts,bm}

\def\eqref#1{equation~\ref{#1}}

\def\1{\bm{1}}

\DeclareMathAlphabet{\mathsfit}{\encodingdefault}{\sfdefault}{m}{sl}
\SetMathAlphabet{\mathsfit}{bold}{\encodingdefault}{\sfdefault}{bx}{n}

\DeclareMathOperator*{\argmax}{arg\,max}

\usepackage{graphicx}  
\usepackage{hyperref}
\usepackage{url}
\usepackage{bm}
\usepackage{multicol}
\usepackage{subfig}
\usepackage{xspace}
\usepackage{xcolor}
\usepackage[export]{adjustbox}
\usepackage{array}
\usepackage{enumitem}

\usepackage{floatrow}
\newfloatcommand{capbtabbox}{table}[][\FBwidth]
\usepackage{blindtext}

\usepackage{multirow}

\newcommand{\srrn}{\textsc{sr-RNN}\xspace}
\newcommand{\srrns}{\textsc{sr-RNN}s\xspace}

\newcommand{\lstm}{\textsc{LSTM}\xspace}
\newcommand{\lstmp}{\textsc{LSTM-p}\xspace}
\newcommand{\lstms}{\textsc{LSTM}s\xspace}
\newcommand{\gru}{\textsc{GRU}\xspace}
\newcommand{\grus}{\textsc{GRU}s\xspace}

\newcommand{\srlstm}{\textsc{sr-LSTM}\xspace}

\newcommand{\srlstmp}{\textsc{sr-LSTM-p}\xspace}
\newcommand{\srlstmps}{\textsc{sr-LSTM-p}s\xspace}

\newcommand{\srgru}{\textsc{sr-GRU}\xspace}
\newcommand{\srgrus}{\textsc{sr-GRU}s\xspace}

\newcommand{\bfb}{\mathbf{b}}
\newcommand{\bfh}{\mathbf{h}}
\newcommand{\bfc}{\mathbf{c}}
\newcommand{\bfx}{\mathbf{x}}
\newcommand{\bfu}{\mathbf{u}}
\newcommand{\bfs}{\mathbf{s}}
\newcommand{\bfp}{\mathbf{p}}
\newcommand{\bff}{\mathbf{f}}
\newcommand{\bfi}{\mathbf{i}}
\newcommand{\bfo}{\mathbf{o}}
\newcommand{\bfy}{\mathbf{y}}
\newcommand{\bfr}{\mathbf{r}}
\newcommand{\bfz}{\mathbf{z}}

\newcommand{\bfW}{\mathbf{W}}
\newcommand{\bfU}{\mathbf{U}}
\newcommand{\bfS}{\mathbf{S}}
\newcommand{\bfR}{\mathbf{R}}

\newcommand{\calQ}{\mathcal{Q}}

\newcommand{\mbc}{\mathbf{s}}
\newcommand{\mbx}{\mathbf{x}}

\usepackage{amsmath, amssymb, amsthm}
\usepackage{algorithm}
\usepackage{algorithmic}
\usepackage{stfloats}



\newtheorem{theorem}{Theorem}[section]

\definecolor{ashgrey}{rgb}{0.5, 0.5, 0.5}
\definecolor{ceruleanblue}{rgb}{0.0, 0.0, 0.0}

\usepackage{color}

\usepackage{graphicx}
\usepackage[colorinlistoftodos]{todonotes}

\usepackage{makecell}

\usepackage{microtype}
\usepackage{graphicx}
\usepackage{booktabs} 

\usepackage{hyperref}

\begin{document}
%
\title{State-Regularized Recurrent Neural Networks to Extract Automata and Explain Predictions}
%
%
%
%

\author{Cheng~Wang\textsuperscript{$\dagger$},~\IEEEmembership{Member ~IEEE,}
	 Carolin~Lawrence, 
         Mathias~Niepert
\IEEEcompsocitemizethanks{
\IEEEcompsocthanksitem \textsuperscript{$\dagger$}Work done while at NEC Laboratories Europe.
\IEEEcompsocthanksitem  Cheng Wang is with Amazon, Berlin, Germany.  E-mail: dr.rer.nat.chengwang@gmail.com.
\IEEEcompsocthanksitem  Carolin Lawrence is with NEC Laboratories Europe, Heidelberg, Germany, E-mail: carolin.lawrence@neclab.eu.
\IEEEcompsocthanksitem  Mathias Niepert is with NEC Laboratories Europe, Heidelberg, Germany and the University of Stuttgart, Stuttgart, Germany, E-mail: mathias.niepert@neclab.eu.
\IEEEcompsocthanksitem  Correspondence author: Cheng Wang.

\protect
}
\thanks{Manuscript received November 1, 2020; revised November 11, 2022.}}

\markboth{IEEE Transactions on Pattern Analysis and Machine Intelligence}%
{Shell \MakeLowercase{\textit{et al.}}: Bare Advanced Demo of IEEEtran.cls for IEEE Computer Society Journals}
%



\IEEEtitleabstractindextext{%
\begin{abstract}
Recurrent neural networks are a widely used class of neural architectures.  They have, however, two shortcomings. First, they are often treated as black-box models and as such it is difficult to understand what exactly they learn as well as how they arrive at a particular prediction. Second, they tend to work poorly on sequences requiring long-term memorization, despite having this capacity in principle. We aim to address both shortcomings with a class of recurrent networks that use a stochastic state transition mechanism between cell applications. This mechanism, which we term state-regularization, makes RNNs transition between a finite set of learnable states. We evaluate state-regularized RNNs on (1) regular languages for the purpose of automata extraction; (2) non-regular languages such as balanced parentheses and palindromes where external memory is required; and (3) real-word sequence learning tasks for sentiment analysis, visual object recognition and text categorisation. We show that state-regularization (a) simplifies the extraction of finite state automata that display an RNN's state transition dynamic; (b) forces RNNs to operate more like automata with external memory and less like finite state machines, which potentiality leads to a more structural memory; (c) leads to better interpretability and explainability of RNNs by leveraging the probabilistic finite state transition mechanism over time steps.
\end{abstract}

}

\maketitle

\IEEEdisplaynontitleabstractindextext

%
\IEEEpeerreviewmaketitle

\ifCLASSOPTIONcompsoc
\IEEEraisesectionheading{\section{Introduction}\label{sec:introduction}}

\fi

%
%
%
%
\IEEEPARstart{R}{ecurrent} neural networks (RNNs) have found their way into numerous applications. Still, RNNs have two shortcomings. 
First, it is difficult to understand what concretely RNNs learn. However, some applications require a close inspection of learned models before deployment and RNNs are more difficult to interpret than rule-based systems. There are a number of approaches for extracting deterministic finite automata (DFAs) from trained RNNs~\cite{giles1991second,Wang:2007:nc,pmlr-v80-weiss18a} as a means to analyze their behavior. These methods apply extraction algorithms after training and it remains challenging to determine whether the extracted DFA faithfully models the RNN's state transition behavior. Most extraction methods are rather complex, depend crucially on hyperparameter choices, and tend to be computationally costly. 
Second, RNNs tend to work poorly on input sequences requiring long-term memorization, despite having this ability in principle. 
Indeed, there is a growing body of work providing evidence, both empirically~\cite{Daniluk:2017,bai:2018,trinh2018learning} and theoretically~\cite{arjovsky2016unitary,zilly2017recurrent,miller:2018}, that recurrent networks offer no benefit on longer sequences, at least under certain conditions. Intuitively, RNNs tend to operate more like DFAs with a large number of states, attempting to memorize all the information about the input sequence solely with their hidden states, and less like automata with external memory.

\begin{figure}%
\centering
\includegraphics[width=\textwidth]{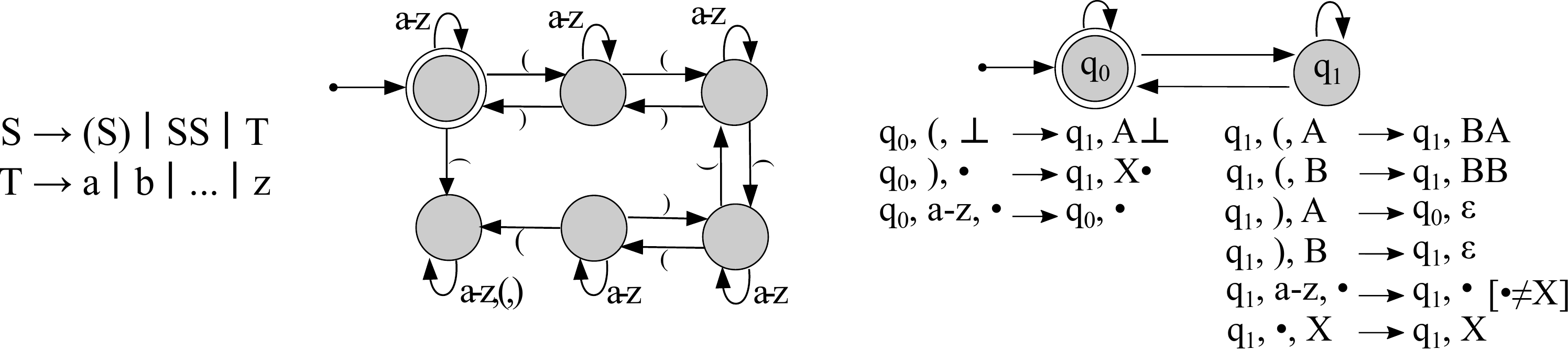} 
\caption{\label{fig:dfa-vs-pda} (Left) The context-free grammar for the language of balanced parentheses (BP). (Center) A DFA that recognizes BP up to depth 4. (Right) A deterministic pushdown automaton (DPDA) that recognizes BP for all depths. The symbol $\bullet$ is a wildcard and stands for all possible tokens. The DPDA extrapolates to all sequences of BP, the DFA recognizes only those up to nesting depth 4.} 
\end{figure}

We propose state-regularized RNNs as a possible step towards addressing both of the aforementioned problems. State-regularized RNNs (\srrns) are a class of recurrent networks that utilize a stochastic state transition mechanism between cell applications.  The stochastic mechanism models a probabilistic state dynamic that lets the \srrns transition between a finite number of learnable states. The parameters of the stochastic mechanism are trained jointly with the parameters of the base RNN. 

\srrns have several advantages over standard RNNs.  First, instead of having to apply post-training DFA extraction, \srrns determine their (probabilistic and deterministic) state transition behavior more directly. We propose a method that extracts DFAs truly representing the state transition behavior of the underlying RNNs. 
Second, we hypothesize that the frequently-observed poor extrapolation behavior of RNNs is caused by memorization with hidden states. It is known that RNNs -- even those with cell states or external memory -- tend to  memorize mainly with their hidden states and in an unstructured manner~\cite{strobelt2016visual,hao:2018}. We show that the state-regularization mechanism shifts representational power to memory components such as the cell state, resulting in improved extrapolation performance.


We support our hypotheses through experiments both on synthetic and real-world datasets.  We explore the  improvement of the extrapolation capabilities of \srrns and closely investigate their memorization behavior. For state-regularized LSTMs, for instance, we observe that memorization can be shifted entirely from the hidden state to the cell state. For text and visual data, state-regularization provides more intuitive interpretations of the RNNs' behavior. A preliminary version of this work appeared in \cite{wang2019state}

\section{Background}
\label{background}

\srrns add state-regularization to RNNs and therefore we first define RNNs (Section \ref{subsec:rnns}). Furthermore, given an \srrn, we describe how a DFA can be extracted faithfully and thus we describe deterministic finite \& pushdown automata  (Section \ref{subsec:automata}).

\subsection{Recurrent Neural Networks (RNNs)}\label{subsec:rnns}
RNNs are powerful learning machines. Siegelmann and Sontag \cite{siegelmann1992computational, siegelmann1994analog, siegelmann2012neural}, for instance, proved that a variant of Elman-RNNs~\cite{elman1990finding} can simulate a Turing machine. Given an input $x_t$ at $t$-th time step:
\begin{align}
  \label{eq:simple-rnn}
    &\bfh_t = \sigma(\bfW_h x_t+\bfU_h \bfh_{t-1}+b_h), \\
    &\bfy_t = \sigma(\bfW_y x_t+b_y),
\end{align}
where $\bfW, \bfU, b_h, b_y$ are parameters, $\sigma$ is the sigmoid activation function, $\bfh$ is the hidden representation and $\bfy$ the output. The key issue for an RNN is to learn to preserve information over long time steps due to vanishing gradients. To alleviate this, several RNN variations have been proposed. Two popular ones are Long Short-Term Memories (LSTMs) \cite{Hochreiter:1997} and Gated Recurrent Units (GRUs) \cite{Chung:2014}.
An LSTM consists of an input gate $\bfi$, a forget gate $\bff$, a memory cell $\bfc$ and an output gate $\bfo$:
\begin{align}
  \label{eq:lstm}
    &\bff_t = \sigma(\bfW_f x_t+\bfU_f \bfh_{t-1}+b_f), \\
    &\bfi_t = \sigma(\bfW_i x_t+\bfU_i \bfh_{t-1}+b_i), \\
    &\bfo_t = \sigma(\bfW_o x_t+\bfU_o \bfh_{t-1}+b_o), \\
    &\hat{\bfc_t} = \phi(\bfW_c x_t+\bfU_c \bfh_{t-1}+b_c), \\
    &\bfc_t = \bff_t \odot \bfc_{t-1} + \bfi_t\odot\hat{\bfc}_t, \\
    &\bfh_t = \bfo_t \odot \phi(\bfc_t);
\end{align}
a GRU has an update gate $\bfz$ and an reset gate $\bfr$:
\begin{align}
  \label{eq:gru}
    &\bfz_t = \sigma(\bfW_z x_t+\bfU_z \bfh_{t-1}+b_z), \\
    &\bfr_t = \sigma(\bfW_r x_t+\bfU_r \bfh_{t-1}+b_r), \\
    &\hat{\bfh}_t = \phi(\bfW_h x_t+\bfr_t \odot \bfh_{t-1}+b_h, \\
    &\bfh_t = \bfz_t \odot \hat{\bfh}_t + (1-\bfz)\odot\bfh_{t-1},
\end{align}
where $\phi$ is hyperbolic tangent function and $\odot$ is the element-wise multiplication operation.
Recent work considers the more practical situation where RNNs have finite precision and linear computation time in their input length~\cite{Weiss:2018-power}.

\subsection{Deterministic Finite \& Pushdown Automata}\label{subsec:automata}
We provide some background on deterministic finite automata (DFAs) and deterministic pushdown automata (DPDAs) for two reasons. First, one contribution of our work is a method for extracting DFAs from RNNs. Second, the state regularization we propose is intended to make RNNs behave more like DPDAs and less like DFAs by limiting their ability to memorize with hidden states.

A DFA is a state machine that  accepts or rejects sequences of tokens and produces one unique computation path for each input. 
Let $\Sigma^{*}$ be the language over the alphabet $\Sigma$ and let $\epsilon$ be the empty sequence.  
A DFA is a 5-tuple $(\calQ, \Sigma, \delta, q_0, F)$ consisting of
a finite set of states $\calQ$, a finite set of input tokens $\Sigma$ called the input alphabet, a transition functions $\delta : \calQ \times \Sigma   \rightarrow \calQ$, a start state $q_0$ and a set of accept states $F \subseteq \calQ$.
A sequence $w$ is accepted by the DFA if the application of the transition function, starting with $q_0$, leads to an accepting state. 
Figure~\ref{fig:dfa-vs-pda} (center) depicts a DFA for the language of balanced parentheses (BP) up to depth 4. A language is regular if and only if it can be described by a DFA.

A pushdown automata (PDA) is defined as a 7-tuple $(\calQ, \Sigma, \Gamma, \delta, q_0, \perp, F)$ consisting of 
a finite set of states $\calQ$; a finite set of input tokens $\Sigma$ called the input alphabet, a finite set of tokens $\Gamma$ called the stack alphabet, a transition function $\delta \subseteq \calQ \times (\Sigma \cup \epsilon) \times \Gamma \rightarrow \calQ \times \Gamma^{*}$, a start state $q_0$, the initial stack symbol $\perp$, and a set of accepting states $F\subseteq \calQ$. Computations of the PDA are applications of the transition relations. The computation starts in $q_0$ with the initial stack symbol $\perp$ on the stack and sequence $w$ as input. The pushdown automaton accepts $w$ if after reading $w$ the automaton reaches an accepting state. 
Figure~\ref{fig:dfa-vs-pda} (right) depicts a deterministic PDA for the language BP.

\section{Related Work}
\label{rel_work}
Our \srrns and applications relate to four different lines of work. First, we discuss existing works that examine how to extract DFAs from RNNs (Section \ref{subsec:dfarnn}). Second, we look at alternative options for regularizing RNNs (Section \ref{subsec:regrnn}). Third, we describe other RNN extensions that modify the state or add an external memory (Section \ref{subsec:statememory}). Fourth, we list approaches to better understand RNNs and discuss how \srrns can support with this effort (Section \ref{subsec:understanding}).
 
\subsection{Extracting DFAs from RNNs}\label{subsec:dfarnn}
  Extracting DFAs from RNNs goes back to work on first-generation RNNs in the 1990s\cite{giles1991second,zeng1993learning}. These methods perform a clustering of hidden states after the RNNs are trained~\cite{wang2018comparison,frasconi1994approach,giles1991second}. Recent work introduced more sophisticated learning approaches to extract DFAs from LSTMs and GRUs~\cite{pmlr-v80-weiss18a}. The latter methods tend to be more successful in finding DFAs behaving similar to the underlying RNN. In contrast to all existing methods, \srrns learn an explicit set of states which facilitates the extraction of DFAs from memory-less \srrns by modelling exactly their state transition dynamics. 
A different line of work attempt to learn more interpretable RNNs~\cite{foerster2017input}, or rule-based classifiers from RNNs~\cite{murdoch2017automatic}.

\subsection{Regularizations of RNNs}\label{subsec:regrnn}
There is a large body of work on regularization techniques for RNNs. Most of these adapt regularization approaches developed for feed-forward networks to the recurrent setting. Representative instances are dropout regularization~\cite{zaremba2014recurrent}, variational dropout~\cite{gal2016theoretically}, weight-dropped LSTMs~\cite{merity2017regularizing}, Zoneout~\cite{krueger2016zoneout} and noise injection \cite{dieng2018noisin}.
Two approaches that can improve convergence and generalization capabilities are batch normalization~\cite{cooijmans2016recurrent} and weight initialization strategies~\cite{le2015simple} for RNNs. In contrast, the proposed \srrns regularize the number of hidden states to a finite set of states. As a result, LSTMs with state regularization can learn in a more structural manner, which leads to improved generalization.

\subsection{State and External Memory Extensions to RNNs}\label{subsec:statememory}
The work most similar to \srrns are self-clustering RNNs~\cite{zeng1993learning}. These RNNs learn discretized states, that is, binary valued hidden state vectors, and it can be shown that these networks generalize better to longer input sequences. Contrary to self-clustering RNNs, we propose an end-to-end differentiable probabilistic state transition mechanism between cell applications. 

Stochastic RNNs are a class of generative recurrent models for sequence data~\cite{bayer2014learning,Fraccaro:2016,Goyal:2017}. They model uncertainty in the hidden states of an RNN by introducing latent variables. 
In contrast to \srrns, stochastic RNNs do not model probabilistic state transition dynamics. Hence, they do not address the problem of overfitting through hidden state memorization nor can they improve DFA extraction. 

There are proposals for extending RNNs with various types of external memory. Representative examples are the neural Turing machine~\cite{graves2014neural}, improvements thereof~\cite{Graves:2016b}, memory network~\cite{WestonCB14}, associative LSTM~\cite{danihelka2016associative}, and RNNs augmented with neural stacks, queues, and deques~\cite{grefenstette2015learning}. Contrary to these proposals, we do not augment RNNs with differentiable data structures but regularize RNNs to make better use of existing memory components such as the cell state. We hope, however, that differentiable neural computers could benefit from state-regularization. 

\subsection{Understanding RNNs}\label{subsec:understanding}
Approaches for understanding CNNs~\cite{simonyan2013deep,zeiler2014visualizing, zhang2018interpretable} have been explored extensively. Studies for interpreting and explaining RNNs are less common.
~\cite{karpathy2015visualizing} revealed the existence of interpretable LSTM cells with character-level language models. ~\cite{li2016visualizing} visualized neural language models. ~\cite{strobelt2016visual} presented a visual analysis tool (namely, LSTMVIS) for visualizing the raw gate activations of LSTMs on understanding these hidden state dynamics over sequences.  While the methods can identify the semantic correlations between hidden cells and abstract attributes or concepts, it is still not obvious how to explain the prediction for given inputs. One of the most recent methods from \cite{murdoch2017automatic} describe a method to extract simple phrase patterns for determining LSTM predictions. With state-regularization we are able to increase the interpretability of RNNs by inspecting the probabilistic state transition over time steps and by directly extracting automata from trained RNN models. 

\begin{figure*}[!htb]
\centering
\includegraphics[width=0.3\textwidth]{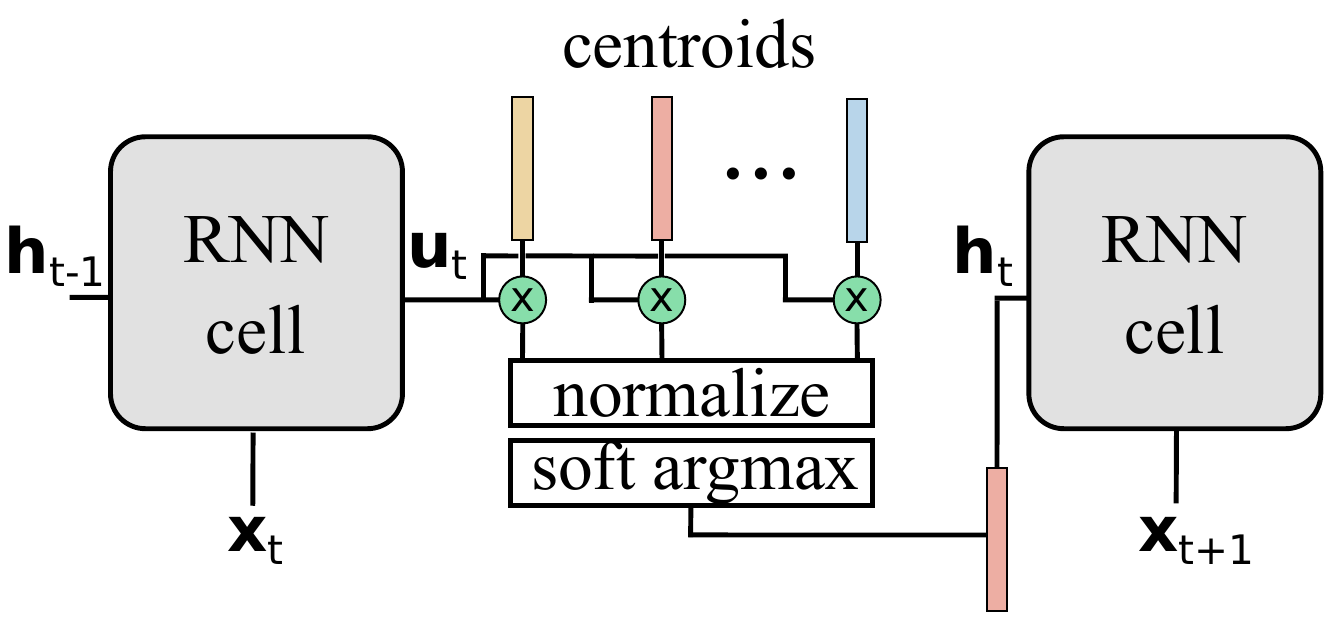}~~~~
\includegraphics[width=0.3\textwidth]{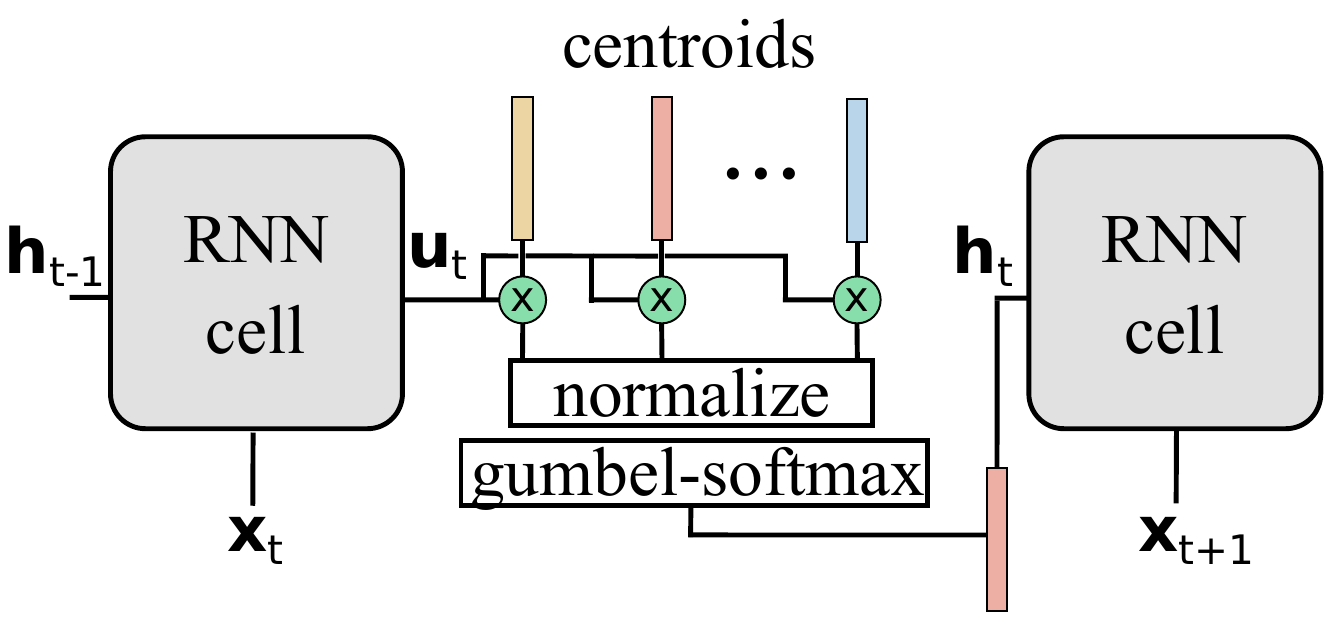}~~~~
\includegraphics[width=0.3\textwidth]{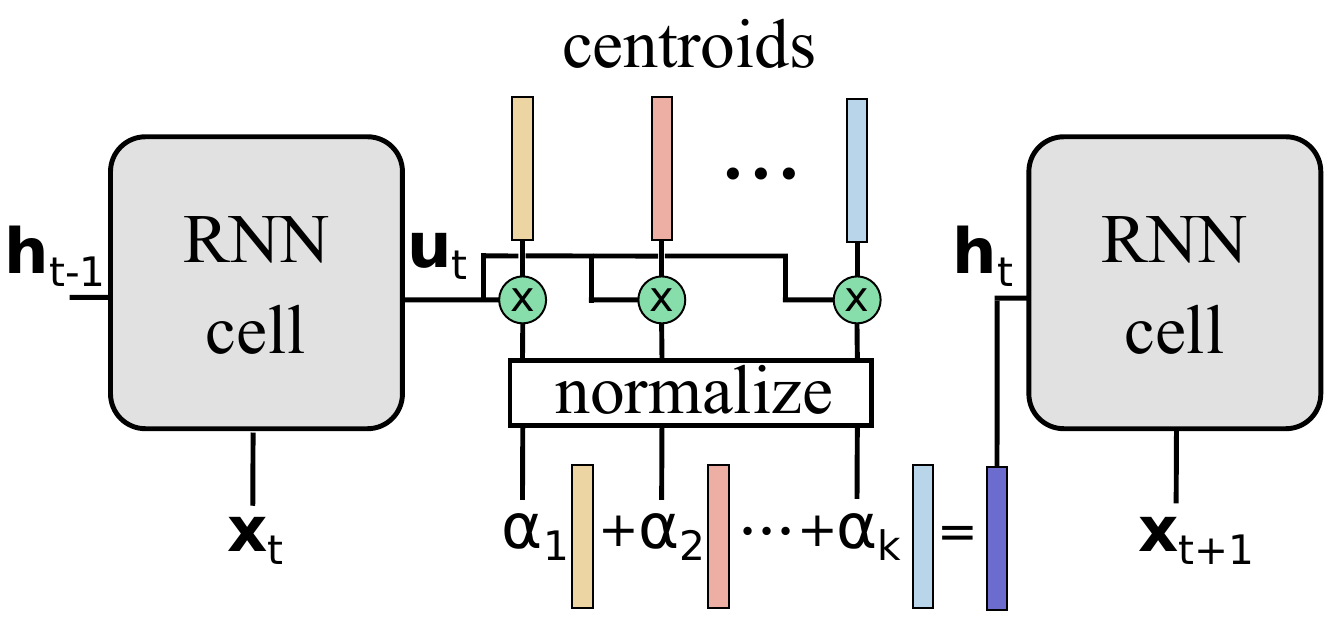}\\
\caption{\label{fig:srnn-example} Three possible instances of an \srrn corresponding to equations \ref{equ:soft_argmax}, \ref{eq:gumbel} and \ref{eqn-transition-2} respectively. }%
\end{figure*}

\section{State-Regularized Recurrent Networks}

The standard recurrence of an RNN is $\bfh_t = f\left(\bfh_{t-1},\bfx_{t}\right)$ where $\bfh_{t-1}$ is the hidden state vector at time $t-1$, and $\bfh_{t}$ and $\bfx_t$ are the hidden state and  the input symbol at time $t$, respectively.  
We refer to RNNs whose unrolled cells are only connected through gated hidden states $\bfh$ as RNNs \emph{without $\infty$-memory}. This is because values of gated hidden states $\bfh$ can only be in a particular interval, such as $[-1, 1]$ for $\mathtt{tanh}$ and not $(-\infty, \infty)$. This limits, in this case, the information flow between cells to values between $-1$ and $1$ and, therefore, memorization has to be performed with fractional changes. The family of \grus is without $\infty$-memory, while \lstms have $\infty$-memory due to their cell state. 

A cell of a state-regularized RNN (\srrn) consist of two components. The first component, which we refer to as the \emph{recurrent component}, applies the function of a standard RNN cell
\begin{equation}
\bfu_{t} = f\left(\bfh_{t-1},\bfc_{t-1},\bfx_{t}\right).
\end{equation}
For the sake of completeness, we include the cell state $\bfc$ here, which is absent in RNNs without $\infty$-memory. 

We propose a second component which we refer to as \emph{stochastic component}. The stochastic component is responsible for modeling the probabilistic state transitions  that let the RNN transition implicitly between a finite number of states. Let $d$ be the size of the hidden state vectors of the recurrent cells. Moreover, let $\Delta^{D} := \{ \bm{\lambda} \in \mathbb{R}_{+}^{D} : \parallel \bm{\lambda} \parallel = 1\}$ be the $(D-1)$ probability simplex.  The stochastic component maintains $k$ learnable centroids $\bfs_1$, ..., $\bfs_k$ of size $d$ which we often write as the column vectors of a matrix $\mathbf{S} \in \mathbb{R}^{d \times k}$. The weights of these centroids are global parameters shared among all cells. The stochastic component computes, at each time step $t$, a discrete probability distribution from the output $\bfu_t$ of the recurrent component  and the centroids of the stochastic component
\begin{equation}
\label{equ:centroids}
\bm{\alpha} = \omega(\mathbf{S}, \bfu_t)  \mbox{ with }  \bm{\alpha} \in \Delta^{k} .  
\end{equation}
Crucially, instances of $\omega$ should be differentiable to facilitate end-to-end training. Typical instances of the function $\omega$ are based on the dot-product,
normalized into a probability distribution
\begin{align}
  \label{eqn-prob-attention}
    \alpha_i = & \frac{ \exp\left((\bfu_{t} \cdot \bfs_i)  / \tau\right) }{ \sum_{i=1}^{k} \exp\left((\bfu_{t} \cdot \bfs_i) / \tau\right)} 
  \end{align}
Here, $\cdot$ is the inner product between two vectors and $\tau$ is a temperature parameter that can be used to anneal the probabilistic state transition behavior. The lower $\tau$ the more $\bm{\alpha}$ resembles the one-hot encoding of a centroid. The higher $\tau$ the more uniform $\bm{\alpha}$ becomes. 
Equation~\ref{eqn-prob-attention} is reminiscent of the equations of attentive mechanisms~\cite{bahdanau2014neural,NIPS2017_7181}. However, instead of attending to the hidden states, \srrns attend to the $k$ centroids to compute transition probabilities. Each $\alpha_i$ is the probability of the RNN to transition to centroid (state) $i$ given the vector $\bfu_t$ for which we write $p_{\bfu_t}(i) =  \alpha_i$. The method has been recently introduced to estimate the uncertainty of RNN~\cite{wang2021uncertainty} and transformer models~\cite{pei2022transformer}.

\begin{figure*}[!htb]
\centering
\subfloat{{\includegraphics[width=0.74\textwidth,valign=b]{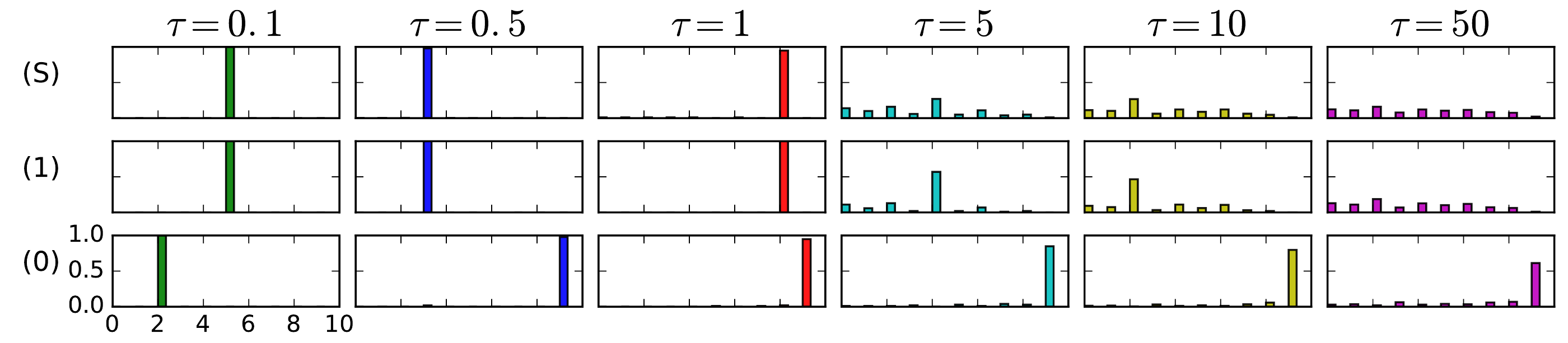} }}%
\hspace{1mm}
\subfloat{{\includegraphics[width=0.17\textwidth,valign=b]{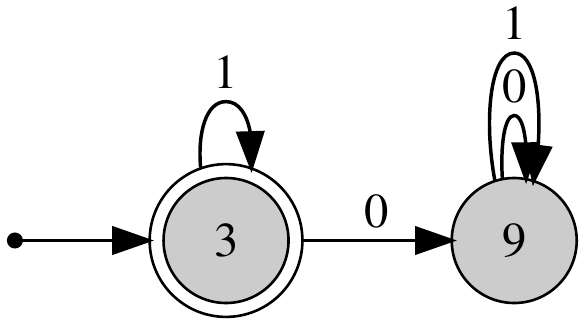} }}%
\vspace{-2mm}
\caption{\label{fig:example-tomita-probs} (Left) State transition probabilities for the \srgru learned from the data for the Tomita~1 grammar, for temperatures $\tau$ and input sequence $\mathtt{[10]}$. $\mathtt{S}$ is the start token. Centroids are listed on x-axis, probabilities on y-axis. Up to temperature $\tau = 1$ the behavior of the trained \srgrus is almost identical to that of a DFA. Despite the availability of $k=10$ centroids, the trained \srgrus use the minimal set of states for $\tau \leq 1$. (Right) The extracted DFA for Tomita grammar 1 and temperature $\tau = 0.5$. }%
\end{figure*}

\subsection{State Transition Mechanisms}

The state transition dynamics of an \srrn is that of a probabilistic finite state machine. At each time step, when in state $\bfh_{t-1}$ and reading input symbol $\bfx_t$, the probability for transitioning to state $\bfs_i$ is $\alpha_i$. Hence, in its second phase, the stochastic component computes the hidden state $\bfh_t$ at time step $t$ from the distribution $\bm{\alpha}$ and the matrix $\bfS$ with a (possibly stochastic) mapping $h: \Delta^{k}\times \mathbb{R}^{d\times k} \rightarrow \mathbb{R}^{d}$. Thus, $\bfh_t = h(\bm{\alpha},\bfS)$. An instance of  $h$ is to 
\begin{equation}
\label{eqn-transition-1}
\mbox{sample } j \sim p_{\bfu_t} \mbox{  and set } \bfh_t = \bfs_j.
\end{equation}
Assigning a particular centroid to be a hidden state, i.e. $\bfh_t = \bfs_j$, is equivalent to an one-hot encoding of centroids. However, the direct application of $\argmax$ renders the \srrn not end-to-end differentiable. To ensure end-to-end differentiability, there are three possible alternative to represent states $\bfs_j$: (1) \textit{soft argmax}, (2) \textit{gumbel-softmax}~\cite{Gumbel:54,JangETAL:17,KendallGal:17} and (3) \textit{mixture of centroids}.

First, a soft and differentiable version of $\argmax$ can be achieved using Equation \ref{eqn-prob-attention} with a low temperature parameter $\tau$ and as $\tau$ approaches 0, $\bm{\alpha}$ approximates a one-hot distribution and is differentiable.
\begin{equation}
\bm{\alpha}= \mbox{\textsc{one\_hot}} \big( \argmax_i ( \frac{ \exp\left((\bfu_{t} \cdot \bfs_i)  / \tau\right) }{ \sum_{j=1}^{k} \exp\left((\bfu_{t} \cdot \bfs_j) / \tau\right)} ) \big),
\label{equ:soft_argmax}
\end{equation}
when $\tau$ approaches 0, the $\bm{\alpha}$ approximates one-hot distribution, but it offers differentiablity to \srrn.

Second, with the gumbel-trick it is possible to draw samples $\bm{z}$ from a categorical distribution given by paramaters $\bm{\theta}$, that is, 
\begin{equation}
\bm{z}= \mbox{\textsc{one\_hot}} \big( \argmax_i [g_i + (\bfu_{t} \cdot \bfs_i)] \big), i \in [1\dots k],
\end{equation}
where $g_i$ are i.i.d. samples from the \textsc{Gumbel}$(0, 1)$, that is, $g=-\log(-\log(u)), u \sim \textsc{uniform}(0, 1)$. Because the $\argmax$ operator breaks end-to-end differentiability, the categorical distribution $\bm{z}$ can be approximated using the differentiable softmax function \cite{JangETAL:17,KendallGal:17}. This enables us to draw a $k$-dimensional sample vector $\bm{\alpha}\in\Delta^{k-1}$, where $\Delta^{k-1}$ is the $(k-1)$-dimensional probability simplex. Each instance $\alpha_i \in \bm{\alpha}$ is assigned a probability, that is, 
\begin{equation}
\label{eq:gumbel}
\alpha_i = \frac{\exp (((\bfu_{t} \cdot \bfs_i)+g_i)/\tau)}{\sum_{j=1}^{k} \exp ((\bfu_{t} \cdot \bfs_j)+g_j)/\tau)}, i \in [1\dots k],
\end{equation}
where $\tau$ is a temperature and $\bm{\alpha}$ approaches $\bm{z}$ as $\tau \rightarrow 0$. Recently~\cite{wang2021uncertainty} showed this approach is able to learn better-calibrated models.

Third, it is possible to set the hidden state $\bfh_t$ to be the probabilistic \textit{mixture of the centroids} 
\begin{equation}
\label{eqn-transition-2}
\bfh_t = \sum_{i=1}^{k} \alpha_i \bfs_i.
\end{equation}
Every internal state $\bfh$ of the \srrn, therefore, is computed as a weighted sum $\bfh = \alpha_1 \bfs_1 + ... + \alpha_k \bfs_k$ of the centroids $\bfs_1, ..., \bfs_k$ with $\bm{\alpha} \in \Delta^{k}$. Here, $h$ is a smoothed variant in contrast to a hard assignment to one of the centroids. 
Figure~\ref{fig:srnn-example} depicts the three variants of the proposed \srrns.


The probabilistic state transition mechanism is also applicable when RNNs have more than one hidden layer. In RNNs with $l>1$ hidden layers, every such layer can maintain its own centroids and stochastic component. In this case, a global state of the \srrn is an $l$-tuple, with the $l{\mbox{th}}$ argument of the tuple corresponding to the centroids of the $l{\mbox{th}}$ layer. 

Even though we have augmented the original RNN with additional learnable parameter vectors, we are actually constraining the \srrn to output hidden state vectors that are similar to the centroids. For lower temperatures and smaller values for $k$, the ability of the \srrn to memorize with its hidden states is increasingly impoverished. We argue that this behavior is beneficial for three reasons:

\begin{itemize}
    \item First, it makes the extraction of interpretable DFAs from memory-less \srrns straight-forward. Instead of applying post-training DFA extraction as in previous work~\cite{Wang:2007:nc,pmlr-v80-weiss18a}, we extract the true underlying DFA directly from the \srrn. Specifically, \srrns don't need an intermediate step that extracts representations from pre-trained RNN models and performs clustering e.g., k-means over the representations. This is automatically done by equations~(\ref{equ:centroids}),(\ref{eqn-prob-attention}),(\ref{eqn-transition-1}) and Algorithm\ref{alg:1}\footnote{It can be integrated to the training loop and output extract DFA at every epoch. Importantly, we don't need to explicity tune the number of clusters $k$.}
    \item Second, we hypothesize that overfitting in the context of RNNs is often caused by memorization via hidden states. Indeed, we show that regularizing the state space pushes representational power to memory components such as the cell state of an \lstm, resulting in improved extrapolation behavior.
    \item Third, the values of hidden states tend to increase in magnitude with the length of the input sequence, a behavior that has been termed \emph{drifting}~\cite{zeng1993learning}. The proposed state regularization stabilizes the hidden states for longer sequences.
\end{itemize}

Next, let us explore some of the theoretical properties of the proposed mechanism. We show that the addition of the stochastic component, when capturing the complete information flow between cells as, for instance, in the case of GRUs, makes the resulting RNN's state transition behavior identical to that of a probabilistic finite state machine. 
\begin{theorem}
\label{theorem-pdfa}
The state transition behavior of an \srrn without $\infty$-memory using Equation~\ref{eqn-transition-1} is identical to that of a probabilistic finite automaton. 
\end{theorem}

\begin{theorem}
\label{theorem-dfa-equiv}
For $\tau \rightarrow 0$ the state transition behavior of an \srrn without $\infty$-memory (using Equations~\ref{eqn-transition-1} or \ref{eqn-transition-2}) is equivalent to that of a deterministic finite automaton. 

\end{theorem}
The proofs of the theorems are part of the appendix. 

\subsection{Learning DFAs with State-Regularized RNNs}
\label{interpretable-rnn}

Extracting DFAs from RNNs is motivated by applications where a thorough understanding of learned neural models is required before deployment. \srrns maintain a set of learnable states and compute and explicitly follow state transition probabilities. It is possible, therefore, to extract finite-state transition functions that truly model the underlying state dynamics of the \srrn. The centroids do not have to be extracted from a clustering of a number of observed hidden states but can be read off of the trained model. This renders the extraction also more efficient. 

\begin{algorithm}[htb]
	\renewcommand{\algorithmicrequire}{\textbf{Input:}}
	\renewcommand{\algorithmicensure}{\textbf{Output:}}
	\caption{Learning transition function (counts-based)}
	\label{alg:1}
	\begin{algorithmic}[1]
		\REQUIRE model $\mathbf{M}$, dataset $\mathbf{D}$, alphabet $\Sigma$, start token $s$
		\ENSURE transition function $\delta$
		\STATE $ \backslash$* \textit{the transition prob. of start token}  *$ \backslash$
		\STATE $\Phi[(\mbc_{i}, \mbx_t, \mbc_j)] = 0$, ~~$i,j  \in \{1, ..., k\},~~ \mbx_t \in \Sigma$
		\STATE $\bm{\alpha} = \mathbf{M}(s), \bm{\alpha} =\{\alpha_i\}, i \in \{1, ..., k\}$
		\STATE $ \backslash$* \textit{select the next state}  *$ \backslash$
		\STATE $i = \argmax_{i \in \{1, ..., k\}} [\bm{\alpha}], ~~\mbc_{start} = \mbc_i $
		\STATE $ \backslash$* \textit{compute and update for each input symbol}  *$ \backslash$
		\FOR{$\bfx = (\mbx_1, \mbx_2,..., \mbx_T) \in \mathbf{D}$}   
     		\FOR{$t \in [1, ..., T]$}  
		\STATE  $ j= \argmax_{i \in \{1, ..., k\}} [\mathbf{M}(\mbx_t)]$, $~~\mbc_{end} = \mbc_j$
		\STATE $\Phi[(\mbc_{start}, \mbx_t, \mbc_{end})] \leftarrow \Phi[(\mbc_{start}, \mbx_t, \mbc_{end})]+ 1$
		\STATE $\mbc_{start}=\mbc_{end}$
		\ENDFOR
		\ENDFOR
		\STATE $ \backslash$* \textit{compute transition function based on transition counts}  *$ \backslash$
		\FOR{$i, j \in \{1, ..., k\} \mbox{ and } \mbx_t \in \Sigma$}   
		\STATE $\delta(\mbc_i, \mbx_t) = \argmax_{j \in \{1, ..., k\}}  \Phi[(\mbc_i, \mbx_t, \mbc_j)]$
		\ENDFOR
	\end{algorithmic}  
\end{algorithm}

We developed two possible approaches, \textit{transition counts}-based and \textit{mean transition probability}-based,  to extract DFAs. First, we adapt previous work~\cite{Schellhammer:1998,Wang:2007:nc} to construct the transition function of an \srrn. We begin with the start token of an input sequence, compute the transition probabilities $\bm{\alpha}$, and move the \srrn to the highest probability state. We continue this process until we have seen the last input token. By doing this, we get a count of transitions from every state $\bfs_i$ and input token $a\in\Sigma$ to the following states (including self-loops).
After obtaining the transition counts, we keep only the most frequent transitions and discard all other transitions. 
Concretely, Algorithm \ref{alg:1} presents the pseudo-code of DFA extraction, where $\Phi[(\mbc_i, \mbx_t, \mbc_j)]$ is a dictionary of transitions and counts, where the tuple $(\mbc_i, \mbx_t, \mbc_j)$ denotes a transition from centroid $\mbc_i$ to $\mbc_j$ given an input symbol $\mbx_t$ and $\delta$ is the returned transition function.

\begin{figure*}[t!]
\vspace{-5mm}
\centering
\subfloat{{\includegraphics[width=0.2\textwidth,valign=b]{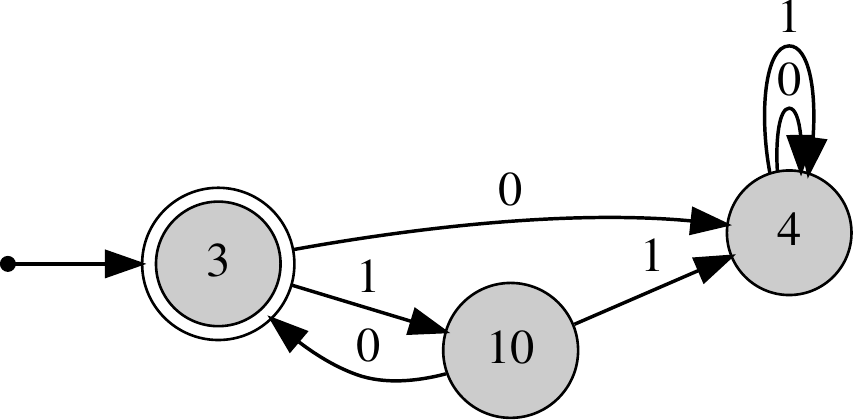} }}%
\hspace{2mm}
\subfloat{{\includegraphics[width=0.28\textwidth,valign=b]{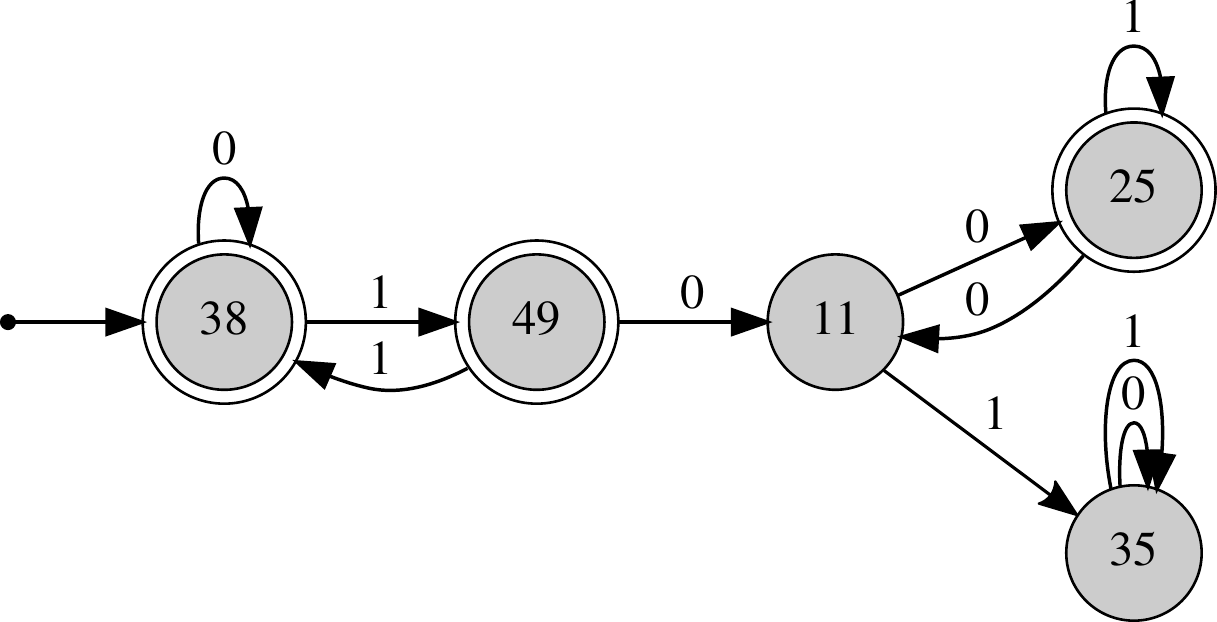} }}%
\hspace{2mm}
\subfloat{{\includegraphics[width=0.28\textwidth,valign=b]{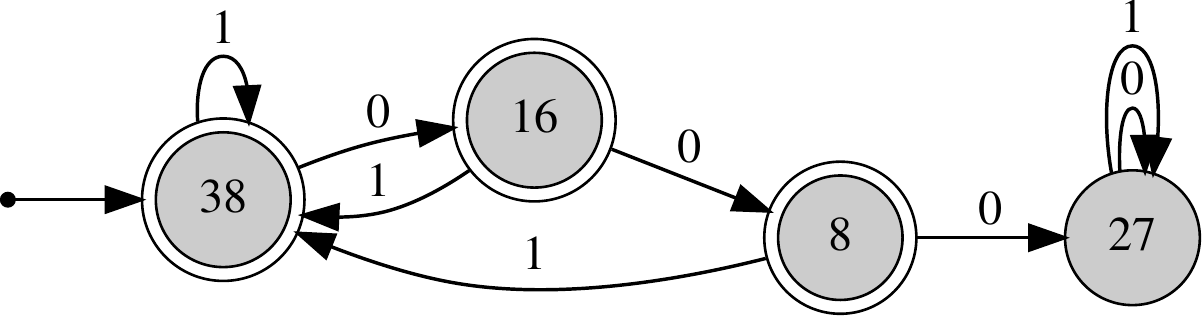} }}%
\caption{\label{fig:example-tomita-dfas} Extracted DFAs of the Tomita grammars 2-4. All DFAs are correct. The state numbers correspond to the index of the learned \srgru centroids.}%
\end{figure*}

 Alternatively, we can learn the transition function based on mean transition probability, rather than transition counts. 
 This can be achieved by computing the mean transition probability:
\begin{equation}
\Phi[(\mbc_{start}, \mbx_t, \mbc_{end})] \leftarrow \Phi[(\mbc_{start}, \mbx_t, \mbc_{end})] +\max [\bm{\alpha}].
\end{equation} 
In this case, $\delta$ is learned by computing the maximum mean transition probability
\begin{equation}
\delta(\mbc_i, \mbx_t) = \argmax_{j \in \{1, ..., k\}}  \mbox{\textsc{Mean}} \left( \Phi[(\mbc_i, \mbx_t, \mbc_j)] \right).
\end{equation}
 
As a corollary of Theorem~\ref{theorem-dfa-equiv}, we have that, for $\tau \rightarrow 0$, the extracted transition function becomes increasingly identical to the transition function of the DFA learned by the \srrn. Figure~\ref{fig:example-tomita-probs} shows that for a wide range of temperatures (including the standard softmax temperature $\tau=1$) the transition behavior of an \srgru is identical to that of a DFA, a behavior we can show to be common when \srrns are trained on regular languages.

\subsection{Learning Non-Regular Languages with State-Regularized LSTMs}
\label{state-regularization}
\begin{table*}
\begin{center}
\begin{small}
\begin{tabular}{|lccc|ccc|}
\hline
Dataset & \multicolumn{3}{c|}{Large Dataset} & \multicolumn{3}{c|}{Small Dataset} \\
Models & \lstm    & \srlstm   & \srlstmp   & \lstm    & \srlstm   & \srlstmp   \\ 
\hline \hline
$d\in{[}1, 10{]}$, $l\leq 100$  & 0.005   & 0.038     & \textbf{0.000}       & 0.068   & 0.037     & \textbf{0.017}       \\
$d\in{[}10, 20{]}$, $l\leq 100$ & 0.334   & 0.255     & \textbf{0.001}       & 0.472   & 0.347     & \textbf{0.189}       \\
$d\in{[}10, 20{]}$, $l\leq 200$ & 0.341   & 0.313     & \textbf{0.003}       & 0.479   & 0.352     & \textbf{0.196}       \\ 
$d=5$,  $l\leq 200$     & 0.002   & 0.044     & \textbf{0.000}       & 0.042   & 0.028     & \textbf{0.015}       \\
$d=10$, $l\leq 200$     & 0.207   & 0.227     & \textbf{0.004}       & 0.409   & 0.279     & \textbf{0.138}       \\
$d=20$, $l\leq 1000$    & 0.543   & 0.540     & \textbf{0.020}       & 0.519   & 0.508     & \textbf{0.380}      \\ 
\hline
\end{tabular}
\end{small}
\end{center}
\caption{\label{tab:bp}Error rates for the balanced parentheses (BP) test sets ($d$=depth, $l$=length, $k$=5 centroids, the training depth $\leq 5$).}
\end{table*}
For more complex languages, such as context-free languages, RNNs that behave like DFAs generalize poorly to longer sequences. The DPDA shown in Figure~\ref{fig:dfa-vs-pda}, for instance, correctly recognizes the language of BP, while the DFA only recognizes it up to nesting depth 4. 
We want to encourage RNNs with memory to behave more like DPDAs and less like DFAs.  
The transition function $\delta$ of a DPDA takes (a) the current state, (b) the current top stack symbol, and (c) the current input symbol and maps these inputs to (1) a new state and (2) a replacement of the top stack symbol (see Section~\ref{background}). Hence, to allow an \srrn, such as the \srlstm, to operate in a manner similar to a DPDA we need to give the RNNs access to these three inputs when deciding what to forget from and what to add to the memory. Precisely this is accomplished for \lstms with peephole connections~\cite{Gers:2000}. 
Concretely, to update the memory, the cell state of a \lstm incorporates forget, input and output gates:
\begin{alignat}{6}
\label{eqn-peephole}
& \textcolor{ashgrey}{\bff_t} & \textcolor{ashgrey}{=} & \ \textcolor{ashgrey}{\sigma \big(\bfW^{f}\bfx_t}  & \textcolor{ashgrey}{+} & \ \textcolor{ashgrey}{\bfR^{f}\bfh_{t-1}}  &  + & \ \textcolor{ceruleanblue}{\bfp^{f} \odot \bfc_{t-1}}  &  \textcolor{ashgrey}{+} & \textcolor{ashgrey}{\bfb^{f}} & \textcolor{ashgrey}{\big)}&, ~~~~~~~~~~~~\\
& \textcolor{ashgrey}{\bfi_t} & \textcolor{ashgrey}{=} & \ \textcolor{ashgrey}{\sigma\big( \bfW^{i}\bfx_t}  & \textcolor{ashgrey}{+} & \ \textcolor{ashgrey}{\bfR^{i}\bfh_{t-1}}  & + & \ \textcolor{ceruleanblue}{\bfp^{i} \odot \bfc_{t-1}}  & \textcolor{ashgrey}{+} & \textcolor{ashgrey}{\bfb^{i}} & \textcolor{ashgrey}{\big)} & ,\\
& \textcolor{ashgrey}{\bfo_t} & \textcolor{ashgrey}{=} & \ \textcolor{ashgrey}{\sigma\big(\bfW^{o}\bfx_t} & \textcolor{ashgrey}{+} & \ \textcolor{ashgrey}{\bfR^{o}\bfh_{t-1}} &  + & \ \textcolor{ceruleanblue}{\bfp^{o} \odot \bfc_{t}}  & \textcolor{ashgrey}{+} & \textcolor{ashgrey}{\bfb^{o}} & \textcolor{ashgrey}{\big)}&,
\end{alignat}
\color{black}
where $\bfh_{t-1}$ is the output of the previous cell's stochastic component; $\bfW$s and $\bfR$s are the matrices of the original LSTM; the $\bfp$s are the parameters of the peephole connections; and $\odot$ is the elementwise multiplication. 
We show empirically that the resulting \srlstmp operates like a DPDA, incorporating the current cell state when making decisions about changes to the next cell state. 

\subsection{Practical Considerations}

Implementing \srrns only requires extending existing RNN cells with a stochastic component. We have found the use of  start and end tokens to be beneficial. The start token is used to transition the \srrn to a centroid representing the start state which then does not have to be fixed a priori. The end token is used to perform one more cell application but without applying the stochastic component before a classification layer. The end token lets the \srrn consider both the cell state and the hidden state to make the accept/reject decision. We find that a temperature of $\tau=1$ (standard softmax) and an initialization of the centroids with values sampled uniformly from $[-0.5,0.5]$ work well across different datasets. 

\subsection{Understanding RNN Models and Predictions}
State-regularization provides new ways to interpret the working of RNNs. Since \srrns have a finite set of states, we can use the observed transition probabilities to visualize their behavior. We argue that the proposed probabilistic state transition mechanism helps to understand RNN in: (1) model interpretation, what the RNN models learned from training data and (2) prediction explanation, the explanation for a specific prediction. We use the concept and definition of interpretation and explaination from Montavona et al.~\cite{montavon2018methods}.

\subsubsection{Extracting RNN Model Prototypes}
To understand what RNN models learn on training data, we are interested in what each learned centroid can represent, because the learning of a centroid is essentially a ``prototype-based clustering'' which is similar to learning vector quantization (LVQ)~\cite{kohonen1995learning}. When RNN models are trained in a supervised manner, the centroids can be learned to represent the semantic meaning of each categorical class. To represent centroids with most representative inputs, we keep track of the input with highest probability for each centroid. The set of words or pixels with highest transition probability are used to represent a specific centroid. We summarize the procedure in Algorithm~\ref{alg:2}.

\begin{algorithm}[htb]
	\renewcommand{\algorithmicrequire}{\textbf{Input:}}
	\renewcommand{\algorithmicensure}{\textbf{Output:}}
	\caption{Generating model prototype}
	\label{alg:2}
	\begin{algorithmic}[1]
		\REQUIRE trained model $\mathbf{M}$, training dataset $\mathbf{D}_{train}$
		\ENSURE the top $N$ prototypical words for each centroid $V_i = \{x\}_i^N$, $i  \in \{1, ..., k\}$ 
		\STATE $\backslash$* \textit{initialise an empty set for each centroid}  *$ \backslash$
		\STATE  $\bfs_i = \{\}$, $i  \in \{1, ..., k\}$
		\FOR{$\mbx = (\mbx_1, \mbx_2,..., \mbx_T) \in \mathbf{D}_{train}$}   
     		\FOR{$t \in [1, ..., T]$}  
     	\STATE $\backslash$* \textit{select the centroid that word $\mbx_t$ has highest prob.}  *$ \backslash$
		\STATE  $ j= \argmax_{i \in \{1, ..., k\}} [\mathbf{M}(\mbx_t)]$
		\STATE $\backslash$* \textit{update the top $N$ words for $j$-th centroid }  *$ \backslash$
		\STATE  $V_j.\textsc{update}(\mbx_t)$
		\ENDFOR
		\ENDFOR
	\end{algorithmic}  
\end{algorithm}

\begin{algorithm}[htb]
	\renewcommand{\algorithmicrequire}{\textbf{Input:}}
	\renewcommand{\algorithmicensure}{\textbf{Output:}}
	\caption{Explaining model prediction}
	\label{alg:3}
	\begin{algorithmic}[1]
		\REQUIRE model $\mathbf{M}$, a test sample $U=\{u_t\}_1^T$, vocabulary $\Sigma$ (in language task)
		\ENSURE model prediction $y$ and its explanation. 
		\STATE $\backslash$* \textit{initialise an empty word-centroid transition matrix $P^i_t$, $i \in \{1, ..., k\}$}  *$ \backslash$
		\STATE $\backslash$* \textit{compute transition prob. for each word $u$ }  *$ \backslash$
		\FOR{$u = (u_1, u_2,..., u_T) \in U$}  
		\STATE $\backslash$* \textit{output $y$, last centroid index $j$, and  $P^i_t$ }  *$ \backslash$
		\STATE $ y, j, P^i_t \Leftarrow  \mathbf{M}(u)$
		\STATE \textit{generate explanations (e.g.,heatmap or highlight words) with $\{u_t\}_1^T, \{P^j_t\}_1^T, \Sigma$} by mapping $P_j^t$ to $u_t$.
		\ENDFOR
	\end{algorithmic}  
\end{algorithm}

\subsubsection{Explaining RNN Predictions}
When applying a trained model to predict on test samples, it is often difficult to understand how the model arrived at the prediction. Probabilistic state transition offers a way to highlight the input symbols which trigger a high transition probability to each centroid. This allows us to highlight the inputs which are highly relevant for the final predictions. We summarize this in Algorithm~\ref{alg:3}.
\begin{table}
\small
\begin{tabular}{|lcccc|}
\hline
Number of centroids & $k=2$   & $k=5$   & $k=10$  & $k=50$   \\ 
\hline \hline
$d\in {[}1, 10{]}$, $l\leq100$  & 0.019 & \textbf{0.017} & 0.021 & 0.034  \\
$d\in {[}10, 20{]}$, $l\leq100$ & \textbf{0.096} & 0.189 & 0.205 & 0.192  \\
$d\in {[}10, 20{]}$, $l\leq200$ & \textbf{0.097} & 0.196 & 0.213 & 0.191  \\  
$d=5$,  $l\leq200$     & 0.014 & 0.015 & \textbf{0.012} & 0.047  \\
$d=10$, $l\leq200$     & \textbf{0.038} & 0.138 & 0.154 & 0.128  \\
$d=20$, $l\leq1000$    & 0.399 & \textbf{0.380} & 0.432 & 0.410 \\  
\hline
\end{tabular}
\caption{\label{tab:bp_k}Error rates of the \srlstmp on the small BP test data for various numbers of centroids $k$ ($d$=depth, $l$=length).}
\end{table}

\begin{figure*}
\vspace{-5mm}
\subfloat[$\bfh_t$ of the \lstm]{{\includegraphics[width=0.275\textwidth,valign=b]{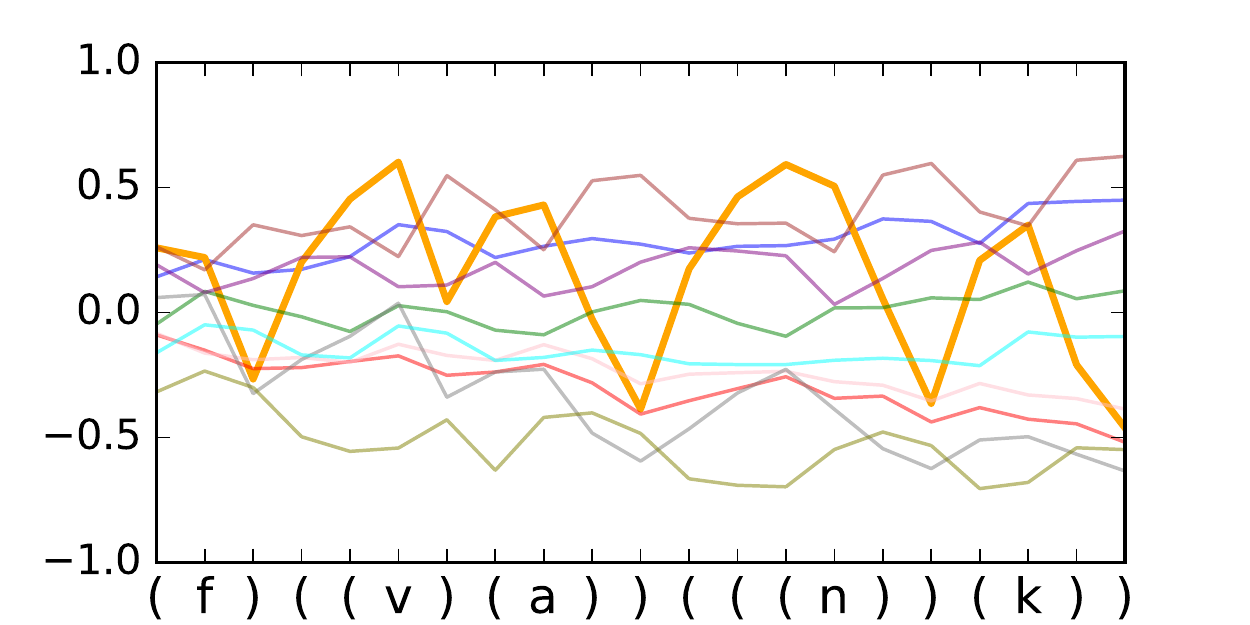} } \hspace{-6.5mm}}%
\subfloat[ $\bfc_t$ of the \lstm]{{\includegraphics[width=0.275\textwidth,valign=b]{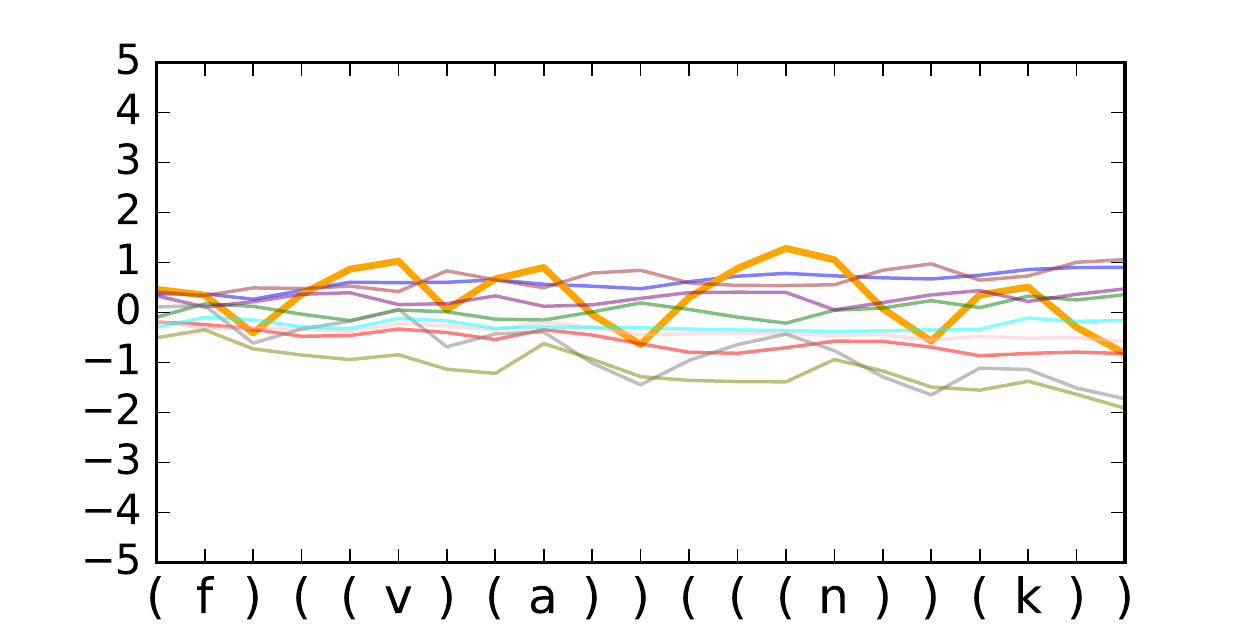} }  \hspace{-6.5mm}} 
\subfloat[ $\bfh_t$ of the \srlstmp]{{\includegraphics[width=0.275\textwidth,valign=b]{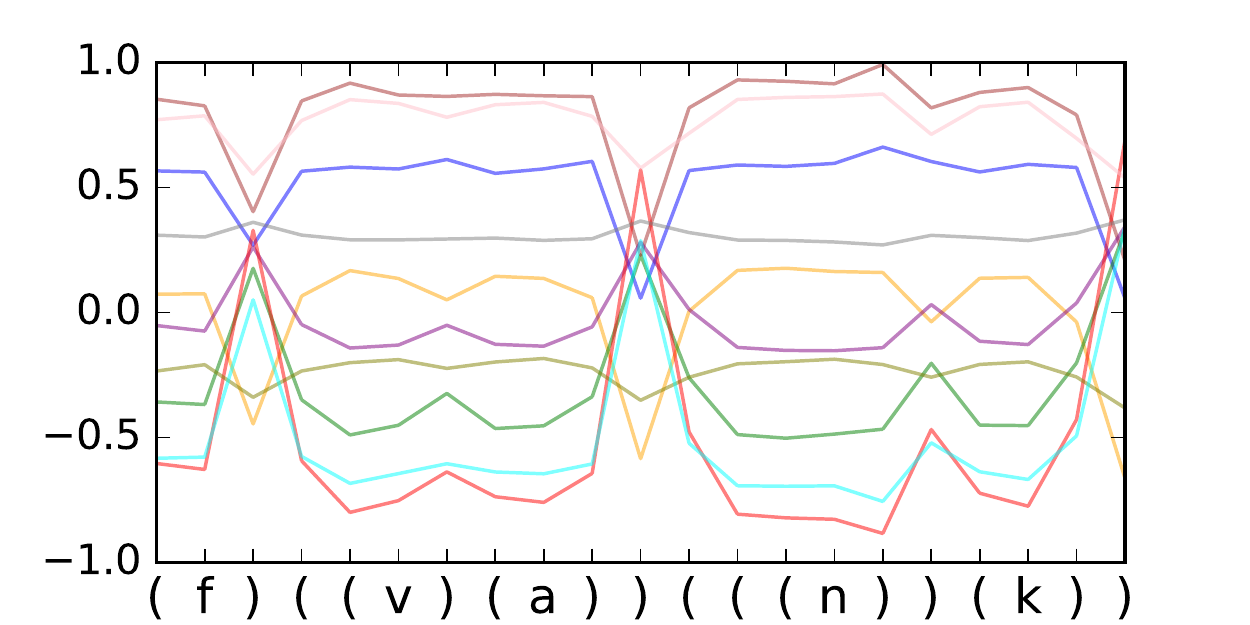} }   \hspace{-6.5mm}}%
\subfloat[ $\bfc_t$ of the \srlstmp]{{\includegraphics[width=0.275\textwidth,valign=b]{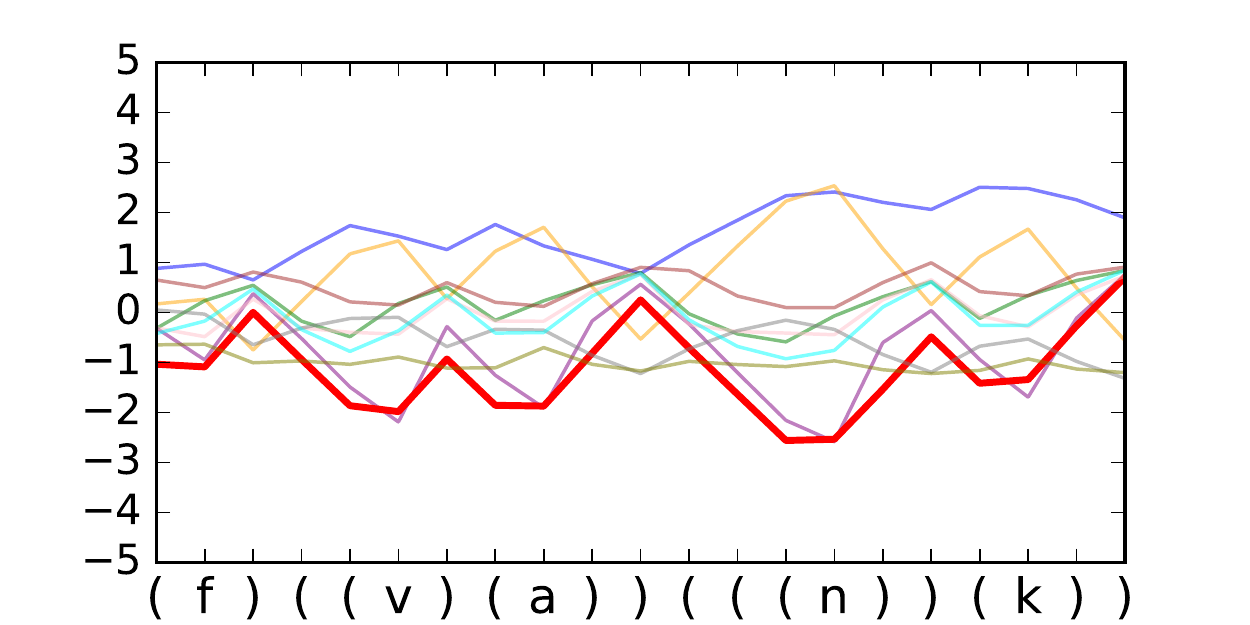} }  \hspace{-6.5mm}}%
\caption{\label{fig:bp_vis} Visualization of hidden state $\bfh_t$ and cell state $\bfc_t$ of the \lstm and the \srlstmp for a specific input sequence from BP.  Each color corresponds to one of 10 hidden units. The \lstm memorizes the number of open parentheses both in the hidden and, to a lesser extent, in the cell state (bold yellow lines). The memorization is not accomplished with saturated gate outputs and a drift is observable for both vectors. The \srlstmp maintains two distinct hidden states (accept and reject) and does not visibly memorize counts through its hidden states. The cell state is used to cleanly memorize the number of open parentheses (bold red line) with saturated gate outputs ($\pm1$). For \srlstmp, a state vector drift is not observable; solutions with less drift generalize better~\cite{gers2001lstm}.}%
\end{figure*}

\section{Experiments}

We conduct four types of experiments to investigate our hypotheses. First, we apply a simple algorithm for extracting DFAs and assess to what extent the true DFAs can be recovered from input data. Second, we compare the behavior of \lstms and state-regularized \lstm on non-regular languages, such as the languages of balanced parentheses and palindromes. Third, we investigate the performance of state-regularized \lstms on non-synthetic datasets. Last, we visualize the probabilistic state transitions to understand RNN models and explain their predictions.

Unless otherwise indicated we always (a) use single-layer RNNs, (b) learn an embedding for input tokens before feeding it to the RNNs, (c) apply \textsc{Adadelta}~\cite{zeiler2012adadelta} for regular language and \textsc{RMSprop}~\cite{tieleman2012lecture} with a learning rate of $0.01$ and momentum of $0.9$ for the rest; (d) do not use dropout or batch normalization of any kind; and (e) use state-regularized RNNs based on Equations~\ref{eqn-prob-attention} and \ref{eqn-transition-2} with a temperature of $\tau=1$ (standard softmax). We implemented \srrns with Theano~\cite{theano}~\footnote{\url{http://www.deeplearning.net/software/theano/}}. 
All experiments were performed on a single Titan Xp with 12G memory. The hyper-parameter were tuned to make sure the vanilla RNNs achieve the best performance.

\subsection{Regular Languages and DFA Extraction}

We evaluate the DFA extraction algorithm for \srrns on RNNs trained on the Tomita grammars~\cite{tomita:cogsci82}, which have been used as
benchmarks in previous  work \cite{Wang:2007:nc,pmlr-v80-weiss18a}. We use available code~\cite{pmlr-v80-weiss18a} to generate training and test data for the regular languages. We first trained a single-layer \gru with $100$ units on the data. We use GRUs since they are memory-less and, hence, Theorem~\ref{theorem-dfa-equiv} applies. Whenever the \gru converged within 1 hour to a training accuracy of $100\%$, we also trained an \srgru based on Equations~\ref{eqn-prob-attention} and \ref{eqn-transition-2} with $k=50$ and $\tau=1$. This was the case for the Grammars 1-4 and 7. The difference in time to convergence between the vanilla \gru and the \srgru was negligible. We applied the transition function extraction outlined in Section~\ref{interpretable-rnn}. In all cases, we could recover the minimal and correct DFA corresponding to the grammars. Figure~\ref{fig:example-tomita-dfas} depicts the DFAs for Grammars 2-4 extracted by our approach. Remarkably, even though we provide more centroids (possible states; here $k=50$) the \srgru only utilizes the required minimal number of states for each of the grammars.  Figure~\ref{fig:example-tomita-probs} visualizes the transition probabilities for different temperatures and $k=10$ for Grammar 1. The numbers on the states  correspond directly to the centroid numbers of the learned \srgru. One can observe that the probabilities are spiky, causing the \srgru to behave like a DFA for $\tau\leq1$.

\subsection{Non-regular Languages}


We conducted experiments on non-regular languages where external memorization is required. This allows us to investigate whether \srlstm behave more like DPDAs and, therefore, extrapolate to longer sequences. 
To this end, we used the context-free language ``balanced parentheses" (BP; see Figure~\ref{fig:dfa-vs-pda} (left)) over the alphabet $\Sigma=\{a, ..., z,(,)\}$,  used in previous work~\cite{pmlr-v80-weiss18a}. We created two datasets for BP. A large one with 22,286 training sequences (positive: 13,025; negative: 9,261) and 6,704 validation sequences (positive: 3,582; negative: 3,122). The small dataset consists of 1,008 training sequences (positive: 601; negative: 407), and 268 validation sequences (positive: 142; negative: 126). Both datasets have 1000 test samples. The training sequences have nesting depths $d \in [1, 5]$,  the validation sequences $d \in [6,10]$ and the test sequences $d \in [0,20]$. We trained the \lstm and the \srrns using curriculum learning as in previous work \cite{Zaremba:2014,pmlr-v80-weiss18a} and using the validation error as stopping criterion. We then applied the trained models to unseen sequences.   Table \ref{tab:bp} lists the results on 1,000 test sequences with the respective depths and lengths. The results show that both \srlstm and \srlstmps extrapolate better on longer sequences and sequences with deeper nesting. Moreover, the \srlstmp performs almost perfectly on the large data, indicating that peephole connections are indeed beneficial.

\begin{figure}
\includegraphics[width=0.6\textwidth]{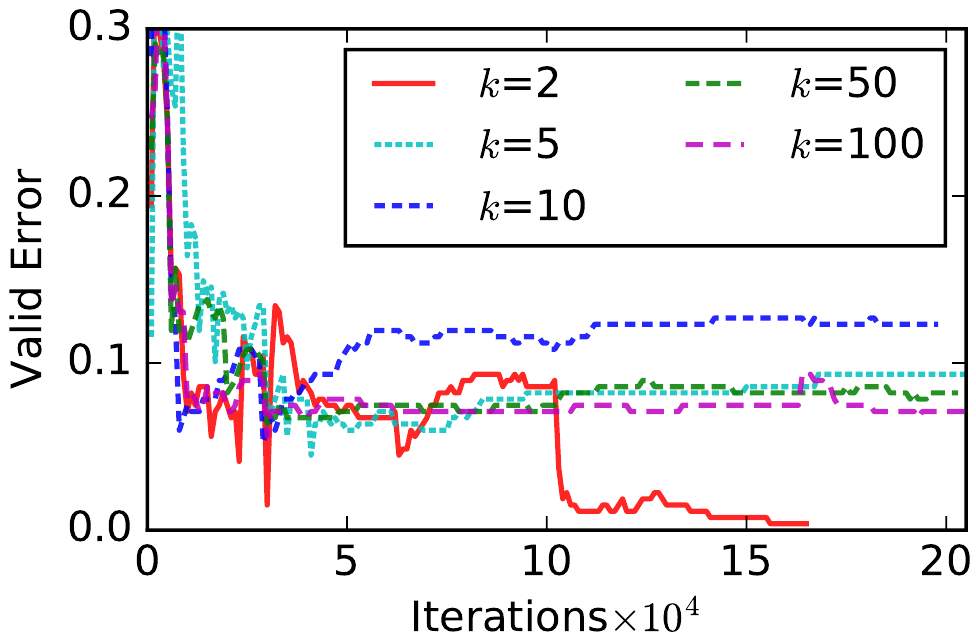}%
 \caption{\label{fig:error-small-bp-k} \srlstmp curves on the small BP validation data. }%
\end{figure}

To explore the effect of the hyperparameter $k$, that is, the number of centroids of the \srrns, we ran experiments on the small BP dataset varying $k$ and keeping everything else the same. Table~\ref{tab:bp_k} lists the error rates and Figure~\ref{fig:error-small-bp-k} the error curves on the validation data for the \srlstmp and different values of $k$. While two centroids ($k=2$) result in the best error rates for most sequence types, the differences are not very pronounced. This indicates that the \srlstmp is robust to changes in the hyperparameter $k$. A close inspection of the transition probabilities reveals that the \srlstmp mostly utilizes two states, independent of the value of $k$. These two states are used as accept and reject states. 
These results show that \srrns generalize and tend to utilize a minimal set states similar to DPDAs.

A major hypothesis of ours is that the state-regularization encourages RNNs to operate more like DPDAs. To explore this hypothesis, we trained an \srlstmp with $10$ units on the BP data and visualized both the hidden state $\bfh_t$ and the cell state $\bfc_t$ for various input sequences. Similar state visualizations have been used in previous work~\cite{strobelt2016visual,Weiss:2018-power}.
Figure~\ref{fig:bp_vis} plots the hidden and cell states for a specific input, where each color corresponds to a dimension in the respective state vectors. As hypothesized, the \lstm relies primarily on its hidden states for memorization. The \srlstmp, on the other hand, does not use its hidden states for memorization. Instead it utilizes two main states (accept and reject) and memorizes the nesting depth cleanly in the cell state. The visualization also shows a drifting behavior for the \lstm, in line with observations made for first-generation RNNs~\cite{zeng1993learning}. Drifting is not observable for the \srlstmp. 

\begin{table}
\begin{tabular}{|l|ccc|}
\hline
Max Length  &    100  & 200 & 500 \\ \hline \hline
\lstm &  31.2 & 42.0 & 47.7 \\
\lstmp &  28.4 & 36.2 & 41.5 \\
\srlstm &  28.0 & 36.0 & 44.6 \\
\srlstmp &  \textbf{10.5} & \textbf{16.7} & \textbf{29.8}  \\ 
\hline
\end{tabular}
\caption{\label{tab:palindromes} Error rates in $\%$ on sequences of varying lengths from the Palindrome test set.}%
\end{table}


We also performed experiments for the non-regular language $ww^{-1}$ (Palindromes)~\cite{schmidhuber2002learning} over the alphabet $\Sigma = \{a, ..., z\}$. We follow the same experiment setup as for BP.
Results are presented in Table \ref{tab:palindromes}. This experiment adds evidence for the improved generalization and memorization behavior of state-regularized LSTMs over vanilla LSTMs (with peepholes).

\subsection{Performance on Real-world Tasks}
Next we test our state regularization on real world datasets, namely on the tasks for sentiment analysis (Section \ref{subsubsec:sa}) and digit recognition (Section \ref{subsubsec:mnist}).

\subsubsection{Sentiment Analysis}\label{subsubsec:sa}

We evaluated state-regularized LSTMs on the IMDB review dataset~\cite{maas-EtAl:2011:ACL-HLT2011}.
It consists of 100k movie reviews (25k training, 25k test, and 50k unlabeled).  We used only the labeled training and test reviews. Each review is labeled as \emph{positive} or \emph{negative}. Table \ref{tab:imdb} lists the results. The \srlstmp is competitive with state of the art methods that also do not use the unlabeled data.
\begin{table}[!htp]
\small
\begin{tabular}{|ll|}
\hline
Methods                                                              & Error \\ \hline \hline
\multicolumn{2}{|c|}{\textbf{use additional unlabeled data}}  \\
Full+unlabelled+BoW~(\cite{maas-EtAl:2011:ACL-HLT2011})   & 11.1         \\
LSTM with tuning and dropout ~(\cite{dai2015semi}) & 13.50 \\
LM-LSTM+unlabelled~(\cite{dai2015semi})              & 7.6            \\
SA-LSTM+unlabelled~(\cite{dai2015semi})           & 7.2             \\  \hline \hline

\multicolumn{2}{|c|}{\textbf{do not use additional unlabeled data}}  \\ 
seq2-bown-CNN~(\cite{johnson2014effective})     & 14.7         \\
Vartional Dropout~(\cite{gal2016theoretically})\footnotemark     & 10.56         \\
WRRBM+BoW(bnc)~(\cite{dahl2012training})     & 10.8          \\
JumpLSTM~(\cite{yu2017learning})       & 10.6         \\ \hline \hline

\lstm & 10.1  \\
\lstmp & 10.3  \\
\srlstm($k=10$) & 9.4  \\
\srlstmp($k=10$, eq. \ref{equ:soft_argmax}) & 11.1 \\
\srlstmp($k=10$, eq. \ref{eq:gumbel}) & 11.0 \\
\srlstmp($k=10$, eq. \ref{eqn-transition-2}) & \textbf{9.2} \\
\srlstmp($k=50$) & 9.8  \\

\bottomrule
\end{tabular}
\caption{\label{tab:imdb} Test error rates (\%) on IMDB.}%
\end{table}

\begin{table}
\small
\begin{tabular}{|lc|} 

\hline
Methods& Error  \\ 
\hline \hline
IRNN~\cite{le2015simple}        & 3.0      \\
URNN~\cite{arjovsky2016unitary}         & 4.9      \\
Full URNN~\cite{NIPS2016_6327}		&2.5		   \\ 
sTANH-RNN ~\cite{zhang2016architectural}   & 1.9 \\
Skip LSTM~\cite{campos2017skip}                    & 2.7    \\
r-LSTM Full BP~\cite{trinh2018learning}                    & 1.6    \\
BN-LSTM~\cite{cooijmans2016recurrent}                  & 1.0    \\
Dilated GRU~\cite{chang2017dilated} & \textbf{0.8}    \\ \hline \hline
\lstm           & 2.3        \\
\lstmp           & 1.5        \\
\srlstm $(k=100)$ &1.4  \\
\srlstmp($k=100$, eq.\ref{equ:soft_argmax}) & 4.1  \\
\srlstmp($k=100$, eq.\ref{eq:gumbel}) & 6.7 \\
\srlstmp ($k=100$, eq. \ref{eqn-transition-2}) &\textbf{0.8}  \\
\srlstmp $(k=50)$ &1.4  \\ \hline
\end{tabular}
\caption{\label{tab:MNIST} Test error (\%) on the sequential MNIST.}%
\end{table}

\begin{table}[htb]
\small
\begin{tabular}{|ll|}
\hline
cent. & words with top-$4$ highest transition probabilities                      \\ \hline\hline
1 &  but (0.97)    hadn (0.91)    college (0.87)    even (0.85)                                    \\
2 &  not (1.0)    or (1.0)    italian (1.0)    never (0.99)                                      \\
3 & loved (1.0)    definitely (1.0)    8 (0.99)    realistic (0.99)    \\
4  &  no (1.0)    worst (1.0)    terrible (1.0)    poorly (1.0)                                         \\ \hline
\end{tabular}
 \caption{\label{tab:imdb_vis_5} The learned centroids and their prototypical words with the top-4 highest transition probabilities on the IMDB dataset. This interprets the \srlstmp model with centroids. The $3^{rd}$ ($4^{th}$) centroid is ``positive'' (``negative'').}
\end{table}

\begin{table*}
\small
\begin{tabular}{|l|lllll|} 
\hline
 concept & \multicolumn{5}{l|}{the top prototypical words and transition probabilities } \\ \hline\hline
talk.politics.misc & comment (0.58)& damage (0.53) & right (0.51) &drug (0.50)  &obligation (0.49) \\ 
talk.politics.mideast & state (0.55) & expansion (0.49) &woman (0.43) &escape (0.35) &nation (0.34) \\ 
comp.graphics & interrupt (0.83) &driver (0.80) &version (0.75) & video (0.55) &network (0.54) \\ 
comp.windows.x & domain (0.48)& information (0.39) &source (0.34) &core (0.31) &message (0.30) \\ \hline
\end{tabular}
\caption{ Interpretation of \srrn ($k=20$ centroids) model trained on the 20NewsGroup. }
\label{tab:20ng_centroids}%
\end{table*}

\subsubsection{Pixel-by-Pixel MNIST}\label{subsubsec:mnist}
We also explored the impact of state-regularization on pixel-by-pixel MNIST~\cite{lecun1998gradient,le2015simple}. Here, the 784 pixels of MNIST images are fed to RNNs one by one  for classification. This requires the ability to memorize long-term dependencies. Table \ref{tab:MNIST} shows the results. The classification function has the final hidden and cell state as input. Our \srlstmps do not use dropout, batch normalization, sophisticated weight-initialization, and are based on a simple single-layer LSTM. We can observe that \srlstmps achieve competitive results, outperforming the vanilla LSTM and LSTM-P. We also conducted additional experiments with state-regularization on vanilla RNNs and GRUs (with the same number of hidden units as \srlstmps). We achieve $31.9$ and $13.6$ test error, respectively, for \srrns and \srgrus, which is worse than the results for the \srlstmps. On MINST both networks failed to converge with same number of training epochs. This suggests the importance of a cell state and $\infty$-memory which we regularized to be used in a more structured manner.

\footnotetext{The number is taken from \cite{wang2021uncertainty}}

\section{Understanding RNNs with Probabilistic State Transitions}
In this section, we visualize the learned centroids and the corresponding transition probabilities to understand the working of RNNs. We use the models trained on IMDB and sequential MNIST from the previous section. Additionally, we train a new \srrn model on 20NewsGroup dataset.\footnote{\url{http://qwone.com/~jason/20Newsgroups/}} The dataset consists of 18,846 samples, among which 15,076 samples are used for training, and 3,770 samples are used for testing. The task is to classify text document into 20 categories. We built a vocabulary with a size of 10,003 and each word is encoded as one-hot representation. \srlstm is able to achieve classification accuracy of 85.6\%. We also trained an \srrn ($k=20$) on a more clean version of the dataset by removing header and footnotes, which achieves an accuracy of 65.7\%.

\subsection{Understanding RNN Models}
Table \ref{tab:imdb_vis_5} lists, for each state (centroid), the word with the top transition probabilities leading to this state. Here the \srrn is trained with $k=5$ centroids. As we can see, only a limited number of centroids are meaningful, for example, the 3$^{rd}$ and 4$^{th}$ represent positive and negative sentiment\footnote{As each centroid $\bfs$ has same dimension as hidden state $\bfh$, we use learned classifier (sofmax layer) to classify centroid so as to decide its categorical label, e.g. positive or negative. If an input word has high transition ($>$0.5) probability to positive centroid, the word is treated as positive as well.}.

\begin{figure}[!htb]
\centering
\includegraphics[width=0.8\textwidth,valign=b]{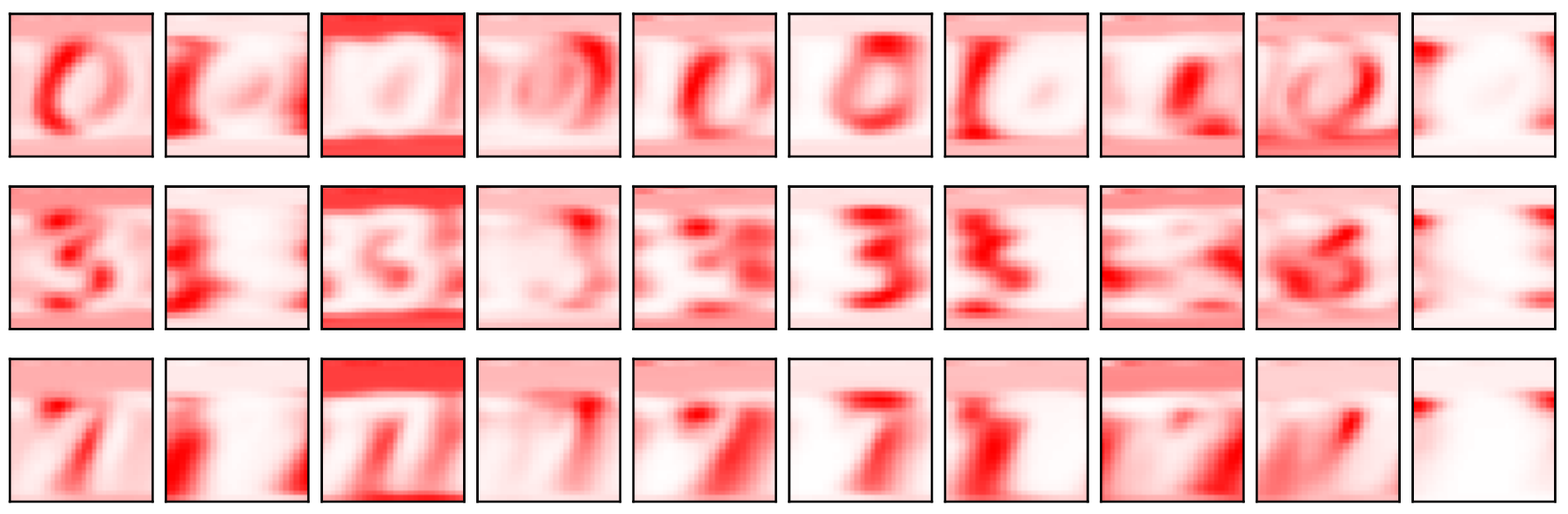}%
\caption{Visualization of the mean transition probability to centroid of \srrns ($k=10$ centroids) models on the Sequential MNIST. The mean transition probabilities can be seen as a ``categorical prototype" for each digit class. Each row represents a digit class and each column depicts the ``categorical prototype" .}%
\label{fig:minst_pro}
\end{figure}

Figure \ref{fig:minst_pro} presents the prototypes for digits 0, 3, 7. Interestingly, we also find the transition functions for each learned centroid are more like the ``kernels" in CNNs. Each centroid intends to capture the different types of features. For example, the $7^{th}$ centroid pays most of its attention to the left part of digits, and the $2^{nd}$ centroid intends to capture the bold property of digits.  

Table \ref{tab:20ng_centroids} shows the prototypical words for some categories in the 20NewsGroup dataset and the corresponding transition probabilities. As shown, the selected words with the highest probabilities are quite sensible (representative) to represent the corresponding class. For example, ``domain, information, source, core, message'' for representing class label ``comp.windows.x''. 
To some extent, this is similar to the topical words generated by topic models (e.g., Latent Dirichlet Allocation (LDA)\cite{blei2003latent}). Differently, \srrns learn the typical words from sequential text data.

\begin{table*}[!htb]
\footnotesize
\begin{tabular}{|p{8.5cm}|p{8.5cm}|} \hline 
 misc.forsale& sci.space \\ \hline\hline 
\colorbox{red!3}{for} \colorbox{red!41}{sale} \colorbox{red!7}{one} \colorbox{red!22}{complete} \colorbox{red!5}{set} \colorbox{red!8}{UNK} \colorbox{red!22}{equipment} \colorbox{red!20}{including} \colorbox{red!5}{base} \colorbox{red!30}{unit} \colorbox{red!19}{portable} \colorbox{red!11}{transmitter} \colorbox{red!6}{UNK} \colorbox{red!18}{plus} \colorbox{red!12}{days} \colorbox{red!11}{free} \colorbox{red!9}{UNK} \colorbox{red!4}{service} \colorbox{red!14}{description} \colorbox{red!8}{item} \colorbox{red!21}{convenient} \colorbox{red!7}{secure} \colorbox{red!4}{anyone} \colorbox{red!3}{whose} \colorbox{red!3}{home} \colorbox{red!14}{broken} \colorbox{red!9}{whose} \colorbox{red!6}{parents} \colorbox{red!15}{live} \colorbox{red!3}{alone} \colorbox{red!3}{children} \colorbox{red!4}{elderly} \colorbox{red!7}{parents} \colorbox{red!5}{UNK} \colorbox{red!3}{heart} \colorbox{red!12}{attack} \colorbox{red!3}{stroke} \colorbox{red!6}{temporarily} \colorbox{red!7}{permanently} \colorbox{red!13}{disabled} \colorbox{red!6}{superior} \colorbox{red!14}{features} \colorbox{red!16}{allows} \colorbox{red!12}{talk} \colorbox{red!7}{UNK} \colorbox{red!9}{center} \colorbox{red!6}{using} \colorbox{red!13}{transmitter} \colorbox{red!2}{help} \colorbox{red!14}{sent} \colorbox{red!5}{soon} \colorbox{red!9}{possible} \colorbox{red!12}{allows} \colorbox{red!16}{personal} \colorbox{red!7}{freedom} \colorbox{red!8}{independence} \colorbox{red!10}{deal} \colorbox{red!7}{item} \colorbox{red!8}{worth} \colorbox{red!4}{us} \colorbox{red!8}{open} \colorbox{red!20}{market} \colorbox{red!33}{asking} \colorbox{red!12}{best} \colorbox{red!47}{offer} \colorbox{red!31}{interested} \colorbox{red!31}{please} \colorbox{red!18}{email} \colorbox{red!20}{UNK} \colorbox{red!17}{UNK} \colorbox{red!17}{stanford} \colorbox{red!29}{edu} \colorbox{red!29}{call} \colorbox{red!18}{will} \colorbox{red!19}{send} \colorbox{red!13}{UNK} \colorbox{red!25}{delivery} \colorbox{red!2}{relevant} \colorbox{red!6}{documents} 

 & 
(...)\colorbox{red!3}{long} \colorbox{red!2}{term} \colorbox{red!18}{planetary} \colorbox{red!2}{monitoring} \colorbox{red!11}{mission} \colorbox{red!1}{occasional} \colorbox{red!1}{chance} \colorbox{red!3}{UNK} \colorbox{red!5}{something} \colorbox{red!7}{like} \colorbox{red!1}{top} \colorbox{red!3}{UNK} \colorbox{red!13}{mission} \colorbox{red!8}{like} \colorbox{red!4}{galileo} \colorbox{red!5}{UNK} 
 \colorbox{red!6}{it} \colorbox{red!5}{unlikely} \colorbox{red!7}{much} \colorbox{red!1}{happening} \colorbox{red!14}{pluto} \colorbox{red!6}{would} \colorbox{red!8}{worth} \colorbox{red!3}{monitoring} \colorbox{red!5}{UNK} \colorbox{red!2}{difficult} \colorbox{red!21}{mission} \colorbox{red!1}{fly} \colorbox{red!5}{without} \colorbox{red!4}{new} \colorbox{red!22}{propulsion} \colorbox{red!12}{technology} \colorbox{red!7}{something} \colorbox{red!29}{planetary} \colorbox{red!0}{community} \colorbox{red!5}{firmly} \colorbox{red!4}{UNK} \colorbox{red!7}{UNK} \colorbox{red!7}{UNK} \colorbox{red!5}{the} \colorbox{red!7}{combined} \colorbox{red!5}{need} \colorbox{red!2}{arrive} \colorbox{red!25}{pluto} \colorbox{red!4}{within} \colorbox{red!2}{reasonable} \colorbox{red!1}{amount} \colorbox{red!3}{time} \colorbox{red!10}{kill} \colorbox{red!1}{nearly} \colorbox{red!1}{cruise} \colorbox{red!1}{velocity} \colorbox{red!0}{settle} \colorbox{red!32}{orbit} \colorbox{red!0}{beyond} \colorbox{red!0}{reasonably} \colorbox{red!2}{done} \colorbox{red!2}{current} \colorbox{red!3}{UNK} \colorbox{red!25}{propulsion} \colorbox{red!17}{most} \colorbox{red!7}{done} \colorbox{red!2}{well} \colorbox{red!22}{earth} \colorbox{red!3}{the} \colorbox{red!19}{things} \colorbox{red!6}{done} \colorbox{red!19}{better} \colorbox{red!13}{voyager} \colorbox{red!18}{like} \colorbox{red!38}{spacecraft} \colorbox{red!13}{UNK} \colorbox{red!21}{need} \colorbox{red!3}{enter} \colorbox{red!38}{orbit} \colorbox{red!31}{around} \colorbox{red!10}{planet} 
\\
\hline
\end{tabular}
\caption{Explaining the \srrn prediction on two 20NewsGroup test samples (stronger highlight indicates higher transition probability). }%
\label{20ng_explain}
\vspace{-2mm}
\end{table*}


\subsection{Explaining RNN Predictions}

\begin{table}[!htb]
\footnotesize
\begin{tabular}{|lp{6.5cm}|}
\hline
Prediction & Test samples \\ \hline \hline
Negative &
\colorbox{red!29}{no}
\colorbox{red!19}{comment}
\colorbox{red!12}{-}
\colorbox{red!24}{stupid}
\colorbox{red!16}{movie}
\colorbox{red!8}{,}
\colorbox{red!1}{acting}
\colorbox{red!27}{average}
\colorbox{red!28}{or}
\colorbox{red!29}{worse}
\colorbox{red!22}{...}
\colorbox{red!16}{screenplay}
\colorbox{red!8}{-}
\colorbox{red!29}{no}
\colorbox{red!26}{sense}
\colorbox{red!25}{at}
\colorbox{red!21}{all}
\colorbox{red!15}{...}
\colorbox{red!25}{skip}
\colorbox{red!21}{it}
\colorbox{red!12}{!} \\ \hline \hline
Positive & 
\colorbox{green!27}{i}
\colorbox{green!6}{thought}
\colorbox{green!21}{this}
\colorbox{green!25}{was}
\colorbox{green!31}{one}
\colorbox{green!32}{of}
\colorbox{green!9}{those}
\colorbox{green!15}{really}
\colorbox{green!30}{great}
\colorbox{green!17}{films}
\colorbox{green!18}{to}
\colorbox{green!11}{see}
\colorbox{green!23}{with}
\colorbox{green!17}{a}
\colorbox{green!5}{bunch}
\colorbox{green!17}{of}
\colorbox{green!3}{close}
\colorbox{green!9}{friends}
\colorbox{green!26}{.}
\colorbox{green!20}{i}
\colorbox{green!35}{laughed}
\colorbox{green!31}{and}
\colorbox{green!9}{cried}
\colorbox{green!28}{and}
\colorbox{green!36}{laughed}
\colorbox{green!32}{and}
\colorbox{green!9}{cried}
\colorbox{green!11}{at}
\colorbox{green!18}{the}
\colorbox{green!29}{same}
\colorbox{green!36}{time} \\ \hline
\end{tabular}
\caption{For a negative (top) and positive (bottom) IMDB prediction, the \srrn highlights the words according to the probability of transitioning to the negative(top)/positive(bottom) centroid (stronger highlight indicates higher transition probability).}
\label{tab:imdb_explain}
\end{table}

\begin{figure}[htb]
\vspace{-2mm}
\centering
\includegraphics[width=0.8\textwidth,valign=b]{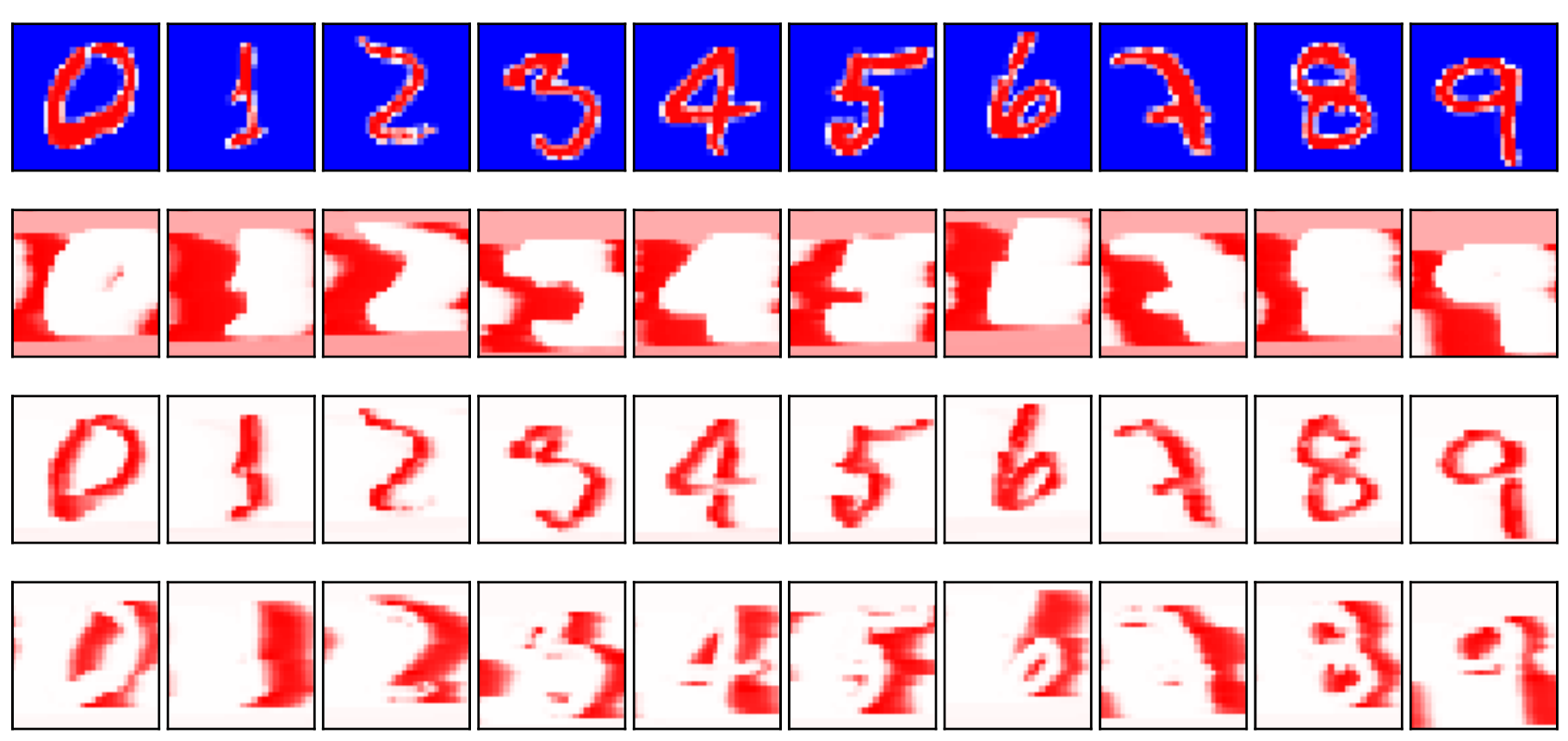}
\caption{For given test samples (top), \srrn ($k=10$) gives the correct predictions on the sequential MNIST. To find the evidences to explain those predictions, we can look at the highest transition probabilities which lead to correct predictions. The 2$^{nd}$ to 4$^{th}$ rows present the highest transition probabilities to the 1$^{st}$ (top), the 3$^{rd}$ (middle) and the 4$^{th}$ centroid respectively.}%
\label{fig:minst_explain}
\end{figure} 

Figure \ref{fig:minst_explain} gives the visual explanations for RNNs predictions on 10 randomly selected MNIST test samples. Table \ref{20ng_explain} presents the explanations of two sample texts from 20NewsGroup which are categorized to ``misc.forsale'' and ``sci.space'' categories. The highlighted words ``sale,  asking, offer delivery'' and ``planetary, orbit, spacecraft, planet'' are highly associative to the RNN predictions.

Table \ref{tab:imdb_explain} presents the predictions of two examples (a positive sample and a negative sample). At each time step, the input words are transitioned between the learned ``positive'' and ``negative'' centroids.  In the negative example, the representative negative words ``no, stupid, worse, skip'' are given high transition probabilities to the negative centroid. 
Similarly, in the bottom example, the words ``great'' and ``laughed''  are associated with the positive centroid. 

\section{Discussion}
We believe that the probabilistic finite state transition has some additional nice properties, which we discuss in the following. 

\textbf{N-gram Phase Extraction}.
With probabilistic finite state transition, we can extend word-level to phrase-level interpretability and explainability. This can be achieved by maintaining an attention window with a size of $n$ on the input sequence. The phrases with the highest mean transition probability can be extracted. Table \ref{tab:phrase_imdb} demonstrates the extracted n-gram phrases for explaining predictions on the IMDB dataset.

\begin{table}[!htb]
\footnotesize
\begin{tabular}{|ll|}
\hline 
2-gram & 4-gram\\ \hline \hline
worst movie &  terrible , terrible ,\\
bad choice &  not a decent performance\\
odd details &  writing : 1 / 10 \\
incredibly awful& movie is extremely boring\\ 
wasted moments & overacting , see it \\ \hline
 great acting    &  10 / 10 . \\
superbly crafted    &    very impressed with this \\
10 !    & movie . great storyline \\
absolutely incredible    &  extremely well composed movie\\
exceptional .    & great cast all around \\ \hline
\end{tabular}
\caption{The n-gram phrases extracted from 1K random IMDB samples for  positive (green) and negative (red) predictions.}
\label{tab:phrase_imdb}
\end{table}

\begin{figure}[!htb]
\centering
\includegraphics[width=0.3\textwidth,valign=b]{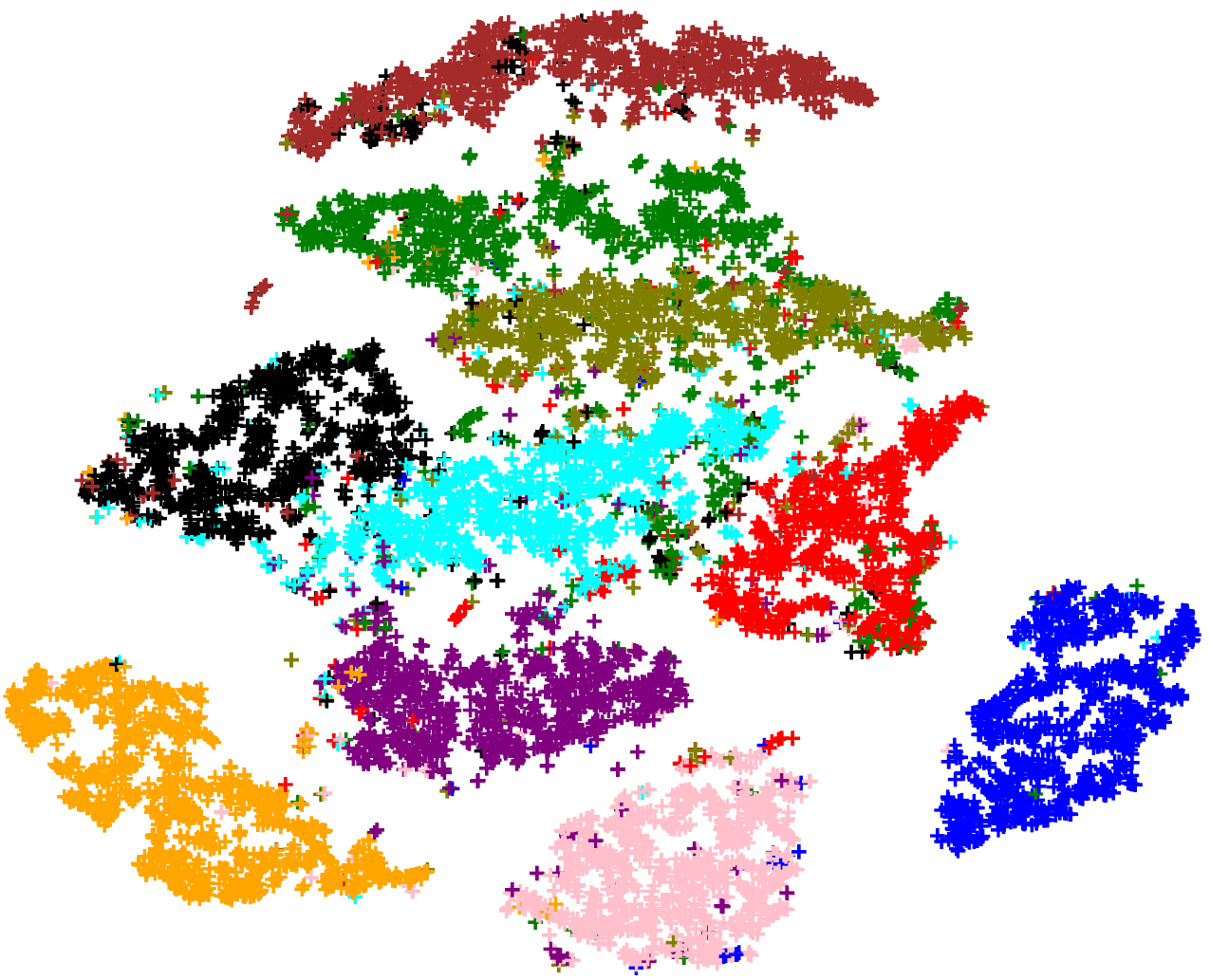}%
\caption{ t-SNE visualization of the extracted finite state transition probabilities for test samples with \srrns ($k=10$) trained on the MNIST.}%
\label{fig:minst_tranistion_feature}
\end{figure}

\textbf{Transitions as Features}. The pattern of probabilistic finite state transition that the \srrn learnt can be used as a representation. Figure \ref{fig:minst_tranistion_feature} shows the t-SNE~\cite{maaten2008visualizing} of finite state transition probabilities for test samples. We find that the transition probabilities of categories are discriminative.

\textbf{Dimensionality Reduction}.
Note that, Figure \ref{fig:minst_tranistion_feature} is not the visualization of $d$-dimension intermediate representation $X \in \mathbb{R}^d$, but the state transition probabilities $P \in \mathbb{R}^k$ of pixel sequences over $k$ centroids. In most cases $k\ll d$ (in this case, $k=10,  d=256$), which also suggests a possibility of using probabilistic finite state transition as dimensionality reduction method.

\section{Conclusion}
State-regularization provides new mechanisms for understanding the workings of RNNs. Inspired by recent DFA  extraction work~\cite{pmlr-v80-weiss18a}, our work simplifies the extraction approach by directly learning a finite set of states and an interpretable state transition dynamic. Even on realistic tasks, such as sentiment analysis, exploring the learned centroids and the transition behavior of \srrns makes for more interpretable RNN models whithout sacrificing accuracy: a single-layer \srrns is competitive with state-of-the-art methods. 
The purpose of our work is not to surpass all existing state of the art methods but to gain a deeper understanding of the dynamics of RNNs. 

State-regularized RNNs operate more like automata with external memory and less like DFAs. This results in a markedly improved extrapolation behavior on several datasets. We do not claim, however, that \srrns are a panacea for all problems associated with RNNs did. For instance, we could not observe an improved convergence of \srrns. Sometimes \srrns converged faster, sometimes vanilla RNNs. While we have mentioned that the computational overhead of \srrns is modest, it still exists, and this might exacerbate the problem that RNNs often take a long to be trained and tuned. We plan to investigate variants of state regularization and the ways in which it could improve differentiable computers with RNN controllers in the future.


%
%
%

\ifCLASSOPTIONcaptionsoff
  \newpage
\fi



%

\bibliography{reference}

\begin{thebibliography}{10}

\bibitem{arjovsky2016unitary}
Martin Arjovsky, Amar Shah, and Yoshua Bengio.
\newblock Unitary evolution recurrent neural networks.
\newblock In {\em ICML}, pages 1120--1128, 2016.

\bibitem{bahdanau2014neural}
Dzmitry Bahdanau, Kyunghyun Cho, and Yoshua Bengio.
\newblock Neural machine translation by jointly learning to align and
  translate.
\newblock {\em ICLR}, 2015.

\bibitem{bai:2018}
Shaojie Bai, J.~Zico Kolter, and Vladlen Koltun.
\newblock An empirical evaluation of generic convolutional and recurrent
  networks for sequence modeling.
\newblock {\em CoRR}, abs/1803.01271, 2018.

\bibitem{bayer2014learning}
Justin Bayer and Christian Osendorfer.
\newblock Learning stochastic recurrent networks.
\newblock {\em arXiv preprint arXiv:1411.7610}, 2014.

\bibitem{blei2003latent}
David~M Blei, Andrew~Y Ng, and Michael~I Jordan.
\newblock Latent dirichlet allocation.
\newblock {\em JMLR}, 2003.

\bibitem{campos2017skip}
V{\'\i}ctor Campos, Brendan Jou, Xavier Gir{\'o}-i Nieto, Jordi Torres, and
  Shih-Fu Chang.
\newblock Skip rnn: Learning to skip state updates in recurrent neural
  networks.
\newblock {\em ICLR}, 2018.

\bibitem{chang2017dilated}
Shiyu Chang, Yang Zhang, Wei Han, Mo~Yu, Xiaoxiao Guo, Wei Tan, Xiaodong Cui,
  Michael Witbrock, Mark~A Hasegawa-Johnson, and Thomas~S Huang.
\newblock Dilated recurrent neural networks.
\newblock In {\em NIPS}, pages 77--87, 2017.

\bibitem{Chung:2014}
Junyoung Chung, Caglar Gulcehre, KyungHyun Cho, and Yoshua Bengio.
\newblock Empirical evaluation of gated recurrent neural networks on sequence
  modeling.
\newblock {\em arXiv preprint arXiv:1412.3555}, 2014.

\bibitem{cooijmans2016recurrent}
Tim Cooijmans, Nicolas Ballas, C{\'e}sar Laurent, {\c{C}}a{\u{g}}lar
  G{\"u}l{\c{c}}ehre, and Aaron Courville.
\newblock Recurrent batch normalization.
\newblock {\em ICLR}, 2017.

\bibitem{dahl2012training}
George~E Dahl, Ryan~P Adams, and Hugo Larochelle.
\newblock Training restricted boltzmann machines on word observations.
\newblock {\em ICML}, 2012.

\bibitem{dai2015semi}
Andrew~M Dai and Quoc~V Le.
\newblock Semi-supervised sequence learning.
\newblock In {\em NIPS}, pages 3079--3087, 2015.

\bibitem{danihelka2016associative}
Ivo Danihelka, Greg Wayne, Benigno Uria, Nal Kalchbrenner, and Alex Graves.
\newblock Associative long short-term memory.
\newblock In {\em ICML}, pages 1986--1994, 2016.

\bibitem{Daniluk:2017}
Michal Daniluk, Tim Rockt{\"a}schel, Johannes Welbl, and Sebastian Riedel.
\newblock Frustratingly short attention spans in neural language modeling.
\newblock 2017.

\bibitem{dieng2018noisin}
Adji~B Dieng, Rajesh Ranganath, Jaan Altosaar, and David~M Blei.
\newblock Noisin: Unbiased regularization for recurrent neural networks.
\newblock {\em ICML}, 2018.

\bibitem{elman1990finding}
Jeffrey~L Elman.
\newblock Finding structure in time.
\newblock {\em Cognitive science}, 14(2):179--211, 1990.

\bibitem{foerster2017input}
Jakob~N Foerster, Justin Gilmer, Jascha Sohl-Dickstein, Jan Chorowski, and
  David Sussillo.
\newblock Input switched affine networks: An rnn architecture designed for
  interpretability.
\newblock In {\em ICML}, pages 1136--1145. JMLR. org, 2017.

\bibitem{Fraccaro:2016}
Marco Fraccaro, S\o ren~Kaae S\o~nderby, Ulrich Paquet, and Ole Winther.
\newblock Sequential neural models with stochastic layers.
\newblock In {\em NeurIPS}, pages 2199--2207. 2016.

\bibitem{frasconi1994approach}
Paolo Frasconi and Yoshua Bengio.
\newblock An em approach to grammatical inference: input/output hmms.
\newblock In {\em ICPR}, pages 289--294. IEEE, 1994.

\bibitem{gal2016theoretically}
Yarin Gal and Zoubin Ghahramani.
\newblock A theoretically grounded application of dropout in recurrent neural
  networks.
\newblock In {\em NeurIPS}, pages 1019--1027, 2016.

\bibitem{gers2001lstm}
Felix~A Gers and E~Schmidhuber.
\newblock Lstm recurrent networks learn simple context-free and
  context-sensitive languages.
\newblock {\em IEEE Transactions on Neural Networks}, 12(6):1333--1340, 2001.

\bibitem{Gers:2000}
Felix~A. Gers and J{\"{u}}rgen Schmidhuber.
\newblock Recurrent nets that time and count.
\newblock In {\em {IJCNN} {(3)}}, pages 189--194, 2000.

\bibitem{giles1991second}
C~Lee Giles, D~Chen, CB~Miller, HH~Chen, GZ~Sun, and YC~Lee.
\newblock Second-order recurrent neural networks for grammatical inference.
\newblock In {\em IJCNN}, volume~2, pages 273--281. IEEE, 1991.

\bibitem{Goyal:2017}
Anirudh Goyal, Alessandro Sordoni, Marc-Alexandre C\^{o}t\'{e}, Nan Ke, and
  Yoshua Bengio.
\newblock Z-forcing: Training stochastic recurrent networks.
\newblock In {\em NeurIPS}, pages 6713--6723. 2017.

\bibitem{graves2014neural}
Alex Graves, Greg Wayne, and Ivo Danihelka.
\newblock Neural turing machines.
\newblock {\em arXiv preprint arXiv:1410.5401}, 2014.

\bibitem{Graves:2016b}
Alex Graves, Greg Wayne, Malcolm Reynolds, Tim Harley, Ivo Danihelka, Agnieszka
  Grabska{-}Barwinska, Sergio~Gomez Colmenarejo, Edward Grefenstette, Tiago
  Ramalho, John Agapiou, Adri{\`{a}}~Puigdom{\`{e}}nech Badia, Karl~Moritz
  Hermann, Yori Zwols, Georg Ostrovski, Adam Cain, Helen King, Christopher
  Summerfield, Phil Blunsom, Koray Kavukcuoglu, and Demis Hassabis.
\newblock Hybrid computing using a neural network with dynamic external memory.
\newblock {\em Nature}, 538(7626):471--476, 2016.

\bibitem{grefenstette2015learning}
Edward Grefenstette, Karl~Moritz Hermann, Mustafa Suleyman, and Phil Blunsom.
\newblock Learning to transduce with unbounded memory.
\newblock In {\em NeurIPS}, pages 1828--1836, 2015.

\bibitem{Gumbel:54}
Emil~Julius Gumbel.
\newblock {Statistical Theory of Extreme Values and Some Practical
  Applications. A Series of Lectures.}
\newblock {\em Number 33. US Govt. Print. Office}, 1954.

\bibitem{hao:2018}
Yiding Hao, William Merrill, Dana Angluin, Robert Frank, Noah Amsel, Andrew
  Benz, and Simon Mendelsohn.
\newblock Context-free transductions with neural stacks.
\newblock In {\em EMNLP workshops}, 2018.

\bibitem{Hochreiter:1997}
Sepp Hochreiter and J\"{u}rgen Schmidhuber.
\newblock Long short-term memory.
\newblock {\em Neural Comput.}, 9(8):1735--1780, 1997.

\bibitem{JangETAL:17}
Eric Jang, Shixiang Gu, and Ben Poole.
\newblock Categorical reparameterization with gumbel-softmax.
\newblock In {\em ICLR, year = {2017}, url =
  {https://openreview.net/forum?id=rkE3y85ee},}.

\bibitem{johnson2014effective}
Rie Johnson and Tong Zhang.
\newblock Effective use of word order for text categorization with
  convolutional neural networks.
\newblock {\em NAACL HLT}, 2015.

\bibitem{karpathy2015visualizing}
Andrej Karpathy, Justin Johnson, and Li~Fei-Fei.
\newblock Visualizing and understanding recurrent networks.
\newblock {\em ICLR workshop}, 2016.

\bibitem{KendallGal:17}
Alex Kendall and Yarin Gal.
\newblock {What Uncertainties Do We Need in Bayesian Deep Learning for Computer
  Vision?}
\newblock In {\em NIPS}, 2017.

\bibitem{kohonen1995learning}
Teuvo Kohonen.
\newblock Learning vector quantization.
\newblock In {\em Self-organizing maps}, pages 175--189. Springer, 1995.

\bibitem{krueger2016zoneout}
David Krueger, Tegan Maharaj, J{\'a}nos Kram{\'a}r, Mohammad Pezeshki, Nicolas
  Ballas, Nan~Rosemary Ke, Anirudh Goyal, Yoshua Bengio, Aaron Courville, and
  Chris Pal.
\newblock Zoneout: Regularizing rnns by randomly preserving hidden activations.
\newblock {\em ICLR}, 2017.

\bibitem{le2015simple}
Quoc~V Le, Navdeep Jaitly, and Geoffrey~E Hinton.
\newblock A simple way to initialize recurrent networks of rectified linear
  units.
\newblock {\em arXiv preprint arXiv:1504.00941}, 2015.

\bibitem{lecun1998gradient}
Yann LeCun, L{\'e}on Bottou, Yoshua Bengio, and Patrick Haffner.
\newblock Gradient-based learning applied to document recognition.
\newblock {\em Proceedings of the IEEE}, 86(11):2278--2324, 1998.

\bibitem{li2016visualizing}
Jiwei Li, Xinlei Chen, Eduard Hovy, and Dan Jurafsky.
\newblock Visualizing and understanding neural models in nlp.
\newblock In {\em NAACL-HLT}, pages 681--691, 2016.

\bibitem{maas-EtAl:2011:ACL-HLT2011}
Andrew~L. Maas, Raymond~E. Daly, Peter~T. Pham, Dan Huang, Andrew~Y. Ng, and
  Christopher Potts.
\newblock Learning word vectors for sentiment analysis.
\newblock In {\em ACL-HLT}, pages 142--150, Portland, Oregon, USA, June 2011.
  Association for Computational Linguistics.

\bibitem{maaten2008visualizing}
Laurens van~der Maaten and Geoffrey Hinton.
\newblock Visualizing data using t-sne.
\newblock {\em JMLR}, 9(Nov):2579--2605, 2008.

\bibitem{merity2017regularizing}
Stephen Merity, Nitish~Shirish Keskar, and Richard Socher.
\newblock Regularizing and optimizing lstm language models.
\newblock {\em ICLR}, 2018.

\bibitem{miller:2018}
John Miller and Moritz Hardt.
\newblock When recurrent models don't need to be recurrent.
\newblock {\em CoRR}, abs/1805.10369, 2018.

\bibitem{montavon2018methods}
Gr{\'e}goire Montavon, Wojciech Samek, and Klaus-Robert M{\"u}ller.
\newblock Methods for interpreting and understanding deep neural networks.
\newblock {\em Digital Signal Processing}, 73:1--15, 2018.

\bibitem{murdoch2017automatic}
W~James Murdoch and Arthur Szlam.
\newblock Automatic rule extraction from long short term memory networks.
\newblock In {\em ICLR}, 2017.

\bibitem{pei2022transformer}
Jiahuan Pei, Cheng Wang, and Gy{\"o}rgy Szarvas.
\newblock Transformer uncertainty estimation with hierarchical stochastic
  attention.
\newblock In {\em Proceedings of the AAAI Conference on Artificial
  Intelligence}, volume~36, pages 11147--11155, 2022.

\bibitem{Schellhammer:1998}
Ingo Schellhammer, Joachim Diederich, Michael Towsey, and Claudia Brugman.
\newblock Knowledge extraction and recurrent neural networks: An analysis of an
  elman network trained on a natural language learning task.
\newblock In {\em Proceedings of the Joint Conferences on New Methods in
  Language Processing and Computational Natural Language Learning}, pages
  73--78, 1998.

\bibitem{schmidhuber2002learning}
J{\"u}rgen Schmidhuber, F~Gers, and Douglas Eck.
\newblock Learning nonregular languages: A comparison of simple recurrent
  networks and lstm.
\newblock {\em Neural Computation}, 14(9):2039--2041, 2002.

\bibitem{siegelmann2012neural}
Hava~T Siegelmann.
\newblock {\em Neural networks and analog computation: beyond the Turing
  limit}.
\newblock Springer Science \& Business Media, 2012.

\bibitem{siegelmann1992computational}
Hava~T Siegelmann and Eduardo~D Sontag.
\newblock On the computational power of neural nets.
\newblock In {\em Proceedings of the fifth annual workshop on Computational
  learning theory}, pages 440--449. ACM, 1992.

\bibitem{siegelmann1994analog}
Hava~T Siegelmann and Eduardo~D Sontag.
\newblock Analog computation via neural networks.
\newblock {\em Theoretical Computer Science}, 131(2):331--360, 1994.

\bibitem{simonyan2013deep}
Karen Simonyan, Andrea Vedaldi, and Andrew Zisserman.
\newblock Deep inside convolutional networks: Visualising image classification
  models and saliency maps.
\newblock {\em arXiv preprint arXiv:1312.6034}, 2013.

\bibitem{strobelt2016visual}
Hendrik Strobelt, Sebastian Gehrmann, Bernd Huber, Hanspeter Pfister,
  Alexander~M Rush, et~al.
\newblock Visual analysis of hidden state dynamics in recurrent neural
  networks.
\newblock {\em CoRR, abs/1606.07461}, 2016.

\bibitem{theano}
{Theano Development Team}.
\newblock {Theano: A {Python} framework for fast computation of mathematical
  expressions}.
\newblock {\em arXiv e-prints}, abs/1605.02688, 2016.

\bibitem{tieleman2012lecture}
Tijmen Tieleman and Geoffrey Hinton.
\newblock Lecture 6.5-rmsprop: Divide the gradient by a running average of its
  recent magnitude.
\newblock {\em COURSERA: Neural networks for machine learning}, 4(2):26--31,
  2012.

\bibitem{tomita:cogsci82}
M.~Tomita.
\newblock Dynamic construction of finite automata from examples using
  hill-climbing.
\newblock In {\em {P}roceedings of the Fourth Annual Conference of the
  Cognitive Science Society}, pages 105--108, 1982.

\bibitem{trinh2018learning}
Trieu~H Trinh, Andrew~M Dai, Thang Luong, and Quoc~V Le.
\newblock Learning longer-term dependencies in rnns with auxiliary losses.
\newblock {\em ICML}, 2018.

\bibitem{NIPS2017_7181}
Ashish Vaswani, Noam Shazeer, Niki Parmar, Jakob Uszkoreit, Llion Jones,
  Aidan~N Gomez, \L~ukasz Kaiser, and Illia Polosukhin.
\newblock Attention is all you need.
\newblock In I.~Guyon, U.~V. Luxburg, S.~Bengio, H.~Wallach, R.~Fergus,
  S.~Vishwanathan, and R.~Garnett, editors, {\em NIPS}, pages 5998--6008. 2017.

\bibitem{wang2021uncertainty}
Cheng Wang, Carolin Lawrence, and Mathias Niepert.
\newblock Uncertainty estimation and calibration with finite-state
  probabilistic {\{}rnn{\}}s.
\newblock In {\em ICLR}, 2021.

\bibitem{wang2019state}
Cheng Wang and Mathias Niepert.
\newblock State-regularized recurrent neural networks.
\newblock In {\em ICML}, pages 6596--6606. PMLR, 2019.

\bibitem{wang2018comparison}
Qinglong Wang, Kaixuan Zhang, II~Ororbia, G~Alexander, Xinyu Xing, Xue Liu, and
  C~Lee Giles.
\newblock A comparison of rule extraction for different recurrent neural
  network models and grammatical complexity.
\newblock {\em arXiv preprint arXiv:1801.05420}, 2018.

\bibitem{Wang:2007:nc}
Qinglong Wang, Kaixuan Zhang, Alexander~G. Ororbia~II, Xinyu Xing, Xue Liu, and
  C.~Lee Giles.
\newblock An empirical evaluation of rule extraction from recurrent neural
  networks.
\newblock {\em Neural Computation}, 30(9):2568--2591, 2018.

\bibitem{pmlr-v80-weiss18a}
Gail Weiss, Yoav Goldberg, and Eran Yahav.
\newblock Extracting automata from recurrent neural networks using queries and
  counterexamples.
\newblock In {\em ICML}, volume~80, pages 5247--5256, 2018.

\bibitem{Weiss:2018-power}
Gail Weiss, Yoav Goldberg, and Eran Yahav.
\newblock On the practical computational power of finite precision rnns for
  language recognition.
\newblock 2018.

\bibitem{WestonCB14}
Jason Weston, Sumit Chopra, and Antoine Bordes.
\newblock Memory networks.
\newblock 2015.

\bibitem{NIPS2016_6327}
Scott Wisdom, Thomas Powers, John Hershey, Jonathan Le~Roux, and Les Atlas.
\newblock Full-capacity unitary recurrent neural networks.
\newblock In D.~D. Lee, M.~Sugiyama, U.~V. Luxburg, I.~Guyon, and R.~Garnett,
  editors, {\em NIPS}, pages 4880--4888. 2016.

\bibitem{yu2017learning}
Adams~Wei Yu, Hongrae Lee, and Quoc~V Le.
\newblock Learning to skim text.
\newblock {\em ACL}, 2017.

\bibitem{Zaremba:2014}
Wojciech Zaremba and Ilya Sutskever.
\newblock Learning to execute.
\newblock {\em CoRR}, abs/1410.4615, 2014.

\bibitem{zaremba2014recurrent}
Wojciech Zaremba, Ilya Sutskever, and Oriol Vinyals.
\newblock Recurrent neural network regularization.
\newblock {\em arXiv preprint arXiv:1409.2329}, 2014.

\bibitem{zeiler2012adadelta}
Matthew~D Zeiler.
\newblock Adadelta: an adaptive learning rate method.
\newblock {\em arXiv preprint arXiv:1212.5701}, 2012.

\bibitem{zeiler2014visualizing}
Matthew~D Zeiler and Rob Fergus.
\newblock Visualizing and understanding convolutional networks.
\newblock In {\em ECCV}, pages 818--833. Springer, 2014.

\bibitem{zeng1993learning}
Zheng Zeng, Rodney~M Goodman, and Padhraic Smyth.
\newblock Learning finite state machines with self-clustering recurrent
  networks.
\newblock {\em Neural Computation}, 5(6):976--990, 1993.

\bibitem{zhang2018interpretable}
Quanshi Zhang, Ying Nian~Wu, and Song-Chun Zhu.
\newblock Interpretable convolutional neural networks.
\newblock In {\em CVPR}, pages 8827--8836, 2018.

\bibitem{zhang2016architectural}
Saizheng Zhang, Yuhuai Wu, Tong Che, Zhouhan Lin, Roland Memisevic, Ruslan~R
  Salakhutdinov, and Yoshua Bengio.
\newblock Architectural complexity measures of recurrent neural networks.
\newblock In {\em NIPS}, pages 1822--1830, 2016.

\bibitem{zilly2017recurrent}
Julian~Georg Zilly, Rupesh~Kumar Srivastava, Jan Koutn{\'\i}k, and J{\"u}rgen
  Schmidhuber.
\newblock Recurrent highway networks.
\newblock In {\em ICML}, pages 4189--4198. JMLR. org, 2017.

\end{thebibliography}

\bibliographystyle{plain}

%

\begin{IEEEbiography}[{\vspace{-1em}\includegraphics[width=1.1in,height=1.2in,clip,keepaspectratio]{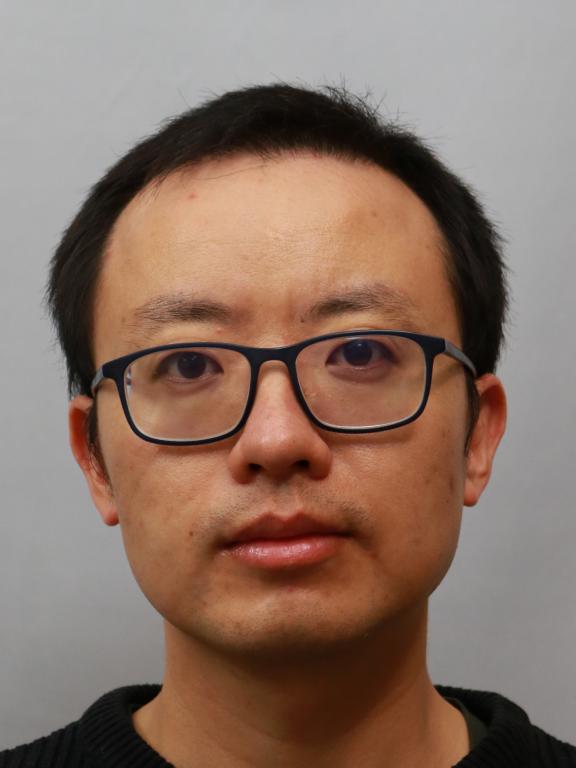}}]
{Cheng Wang} is a machine learning scientist. He received his Dr. rer. nat. degree from Hasso Plattner Institute, the University of Potsdam (2017). His research interests are machine learning, multimodal deep learning and recurrent neural networks with applications in language and vision, information retrieval tasks. He is a PC member of ICML, NeurPS, ICLR, NAACL, ACL, EMNLP, AAAI, IJCAI, ACMMM and an invited reviewer of AIJ, IEEE TNNLS, IEEE TIP, IEEE TKDE, IEEE TMM etc.. He is IEEE and ACM member.  
\end{IEEEbiography}
\vspace{-10mm}
\begin{IEEEbiography}[{\vspace{-1em}\includegraphics[width=1.1in,height=1.2in,clip,keepaspectratio]{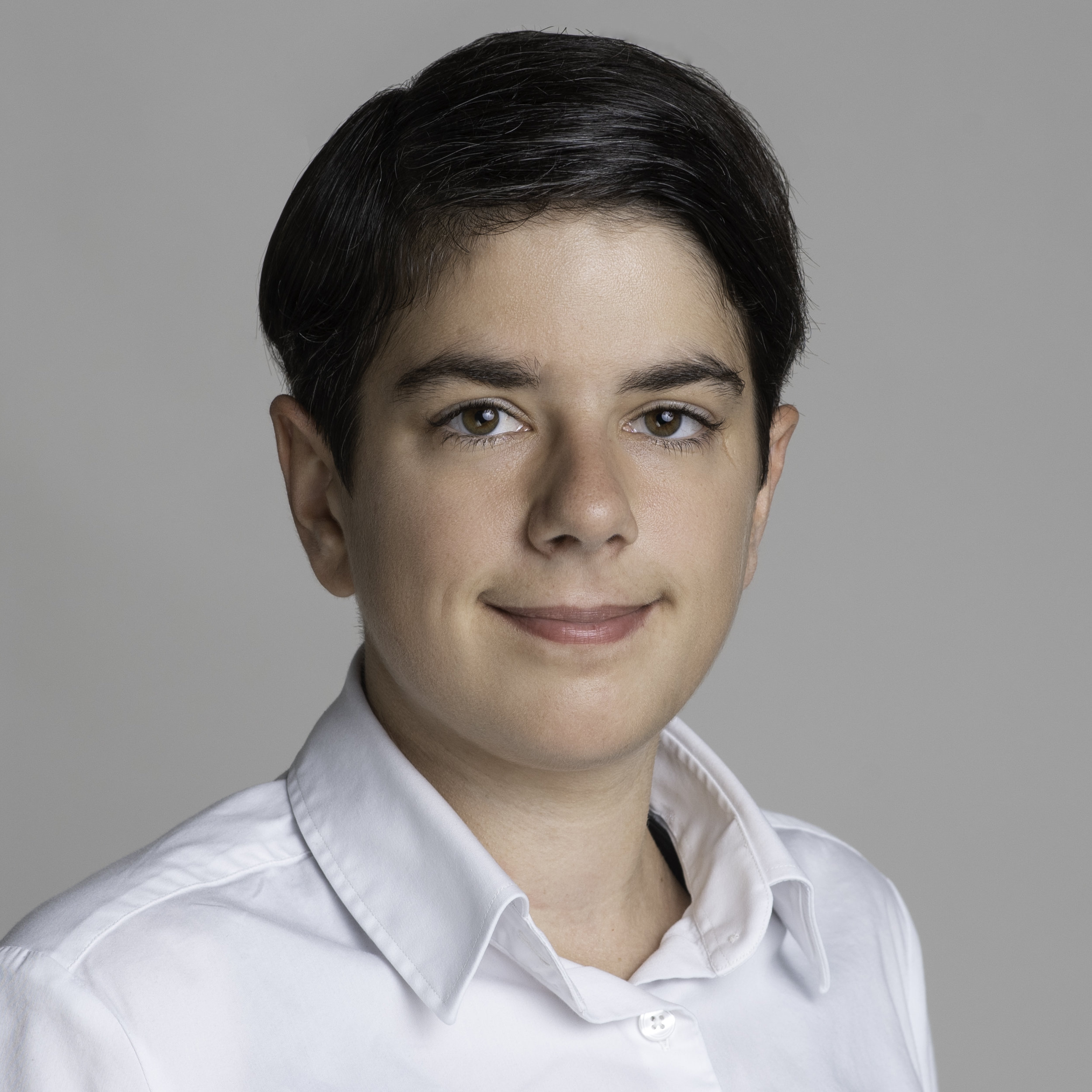}}]
{Carolin Lawrence} is a manager at NEC Laboratories Europe. She received her PhD with the highest distinction in Computational Linguistics from Heidelberg University, Germany (2019). Her research focus includes explainable AI, human-centric AI, NLP and knowledge graphs. She is a senior programm committee member of top-tier NLP conferences (ACL, NAACL, EMNLP). She won the outstanding paper award at the leading conference for knowledge graphs, Automated Knowledge Base Construction (AKBC), in 2021.
\end{IEEEbiography}
\vspace{-10mm}
\begin{IEEEbiography}[{\vspace{-1em}\includegraphics[width=1.1in,height=1.2in,clip,keepaspectratio]{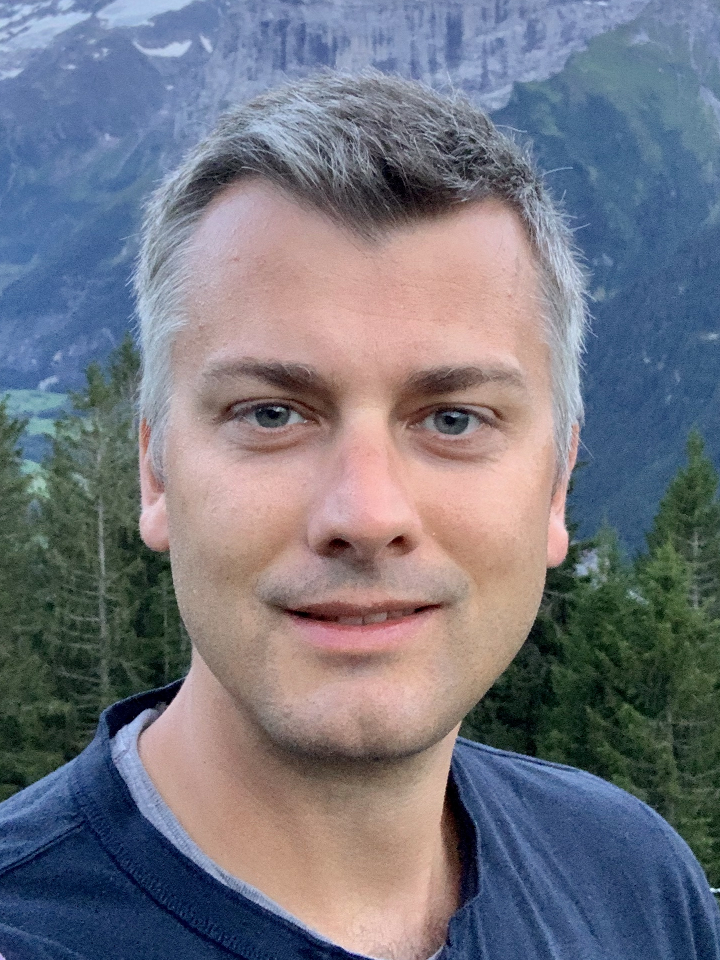}}]
{Mathias Niepert} is a professor at the University of Stuttgart, a faculty member of the Max Planck Research School for Intelligent Systems and ELLIS, and Chief Scientific Advisor at NEC Labs Europe. He received his PhD from Indiana University (2009) and was a postdoctoral researcher at the University of Washington, Seattle. His research interests include deep geometric learning, unsupervised and semi-supervised learning, and probabilistic graphical models. He has won several best paper awards and grants such as a Google Research Award. He is a PC member and/or area chair of top conferences such as ICML, NeurIPS, UAI, ICLR, AAAI and IJCAI.
\end{IEEEbiography}





\clearpage
\pagebreak
\newpage

\begin{figure*}[!b]
\begin{align}
\label{eq:supp}
\alpha_i = & \frac{ \exp\left( b_i  / \tau\right) }{ \exp(\left(b_M-\epsilon_1) / \tau\right) + ... + \exp\left(b_M / \tau\right) + ... + \exp(\left(b_M-\epsilon_k) / \tau\right)} \\
\label{eq:supp1}
& =\frac{ \exp\left( b_i  / \tau\right) }{ \exp(b_M/\tau)\exp(\epsilon_1/ \tau)^{-1} + ... + \exp(b_M / \tau) + ... + \exp(b_M/\tau)\exp(\epsilon_k/\tau)^{-1}}  \\
\label{eq:supp2}
& =\frac{ \exp\left( b_i  / \tau\right) }{ \exp(b_M/\tau)\left[ \exp(\epsilon_1/ \tau)^{-1} + ... + 1 + ... + \exp(\epsilon_k/\tau)^{-1}\right]}.
\end{align}
\end{figure*}


\section*{Supplementary Materials}
\subsection*{Proofs of Theorems \ref{theorem-pdfa} and \ref{theorem-dfa-equiv}}

\noindent \textit{Theorem} \ref{theorem-pdfa}: \textbf{The state transition behavior of an \srrn without $\infty$-memory using equation~\ref{eqn-transition-1} is identical to that of a probabilistic finite automaton. }

\begin{proof}
The state transition function $\delta$ of a probabilistic finite state machine is identical to that of a finite deterministic automaton (see section~\ref{background}) with the exception that it returns a probability distribution over states. For every state $q$ and every input token $a$ the transition mapping $\delta$ returns a probability distribution 
\begin{align}
\bm{\alpha} = (\alpha_1, ..., \alpha_k) = \delta(q, a)
\end{align}
that assigns a fixed probability to each possible state $q \in \mathcal{Q}$ with $|\mathcal{Q}|=k$. The automaton transitions to the next state according to this distribution. Since by assumption the \srrn is using equation~\ref{eqn-transition-1}, we only have to show that the probability distribution over states computed by the stochastic component of an \srrn without $\infty$-memory is identical for every state $q$ and every input token $a$ irrespective of the previous input sequence $\mathbf{a} = \{a_1, ..., a_n\}$ and corresponding state transition history $\mathbf{q} = \{q_{1}, ..., q_{n}\}$. 
\begin{align}
\label{eq:supp_state_trans}
\delta(q_{1}, ..., q_{n}, q, a_1, ..., a_n, a) = \delta(q, a)
\end{align}

More formally, for every pair of input token sequences $\mathbf{a}_1$ and $\mathbf{a}_2$ with corresponding pair of resulting state sequences $\mathbf{q}_1 = (q_{i_1}, ..., q_{i_n}, q)$ and $\mathbf{q}_2 = (q_{j_1}, ..., q_{j_m}, q)$ in \srrn without $\infty$-memory, we have to prove, for every token $a \in \Sigma$, that $\bm{\alpha}_1$ and $\bm{\alpha}_2$, the probability distributions over the states returned by the stochastic component for state $q$ and input token $a$, are  identical. 
\begin{align}
\label{eq:supp_state_trans_1}
\delta(q_{i_1}, ..., q_{i_n}, q, a) = \delta(q_{j_1}, ..., q_{j_m}, q, a)
\end{align}

Now, since the RNN is, by assumption, without $\infty$-memory, we have for both $\mathbf{a}_1, \mathbf{q}_1$ and $\mathbf{a}_2, \mathbf{q}_2$ that the only inputs to the RNN cell are exactly the centroid $\mathbf{s}_{q}$ corresponding to state $q$ and the vector representation of token $a$. Hence, under the assumption that the parameter weights of the RNN are the same for both state sequences $\mathbf{q}_1$ and $\mathbf{q}_2$, we have that the output $\mathbf{u}$ of the recurrent component (the base RNN cell) is identical for $\mathbf{q}_1$ and $\mathbf{q}_2$. Finally, since by assumption the centroids $\mathbf{s}_1, ..., \mathbf{s}_k$ are fixed, we have that the returned probability distributions $\bm{\alpha}_1$ and  $\bm{\alpha}_2$ are identical. Hence, the transition behavior of \srrn without $\infty$-memory is identical to that of a probabilistic finite automaton. 
\end{proof}

\noindent \textit{Theorem} \ref{theorem-dfa-equiv}: \textbf{For $\tau \rightarrow 0$ the state transition behavior of an \srrn without $\infty$-memory (using equations~\ref{eqn-transition-1} or \ref{eqn-transition-2}) is equivalent to that of a deterministic finite automaton.}

\begin{proof}
Let us consider the softmax function with temperature parameter 
$\tau$ $$\alpha_i = \frac{ \exp\left( b_i  / \tau\right) }{ \sum_{i=1}^{k} \exp\left(b_i / \tau\right)}$$ for $1 \leq i \leq k$. \srrns use this softmax function to normalize the scores (from a dot product) into a probability distribution. First, we show that for $\tau \rightarrow 0^{+}$, that there is exactly one $M \in \{1, ..., k\}$ such that $\alpha_M = 1$ and $\alpha_i = 0$ for all $i \in \{1, ..., k\}$ with $i\neq M$.
Without loss of generality, we assume that there is a $M \in \{1, ..., k\}$ such that $b_M > b_i$ for all $i \in \{1, ..., k\}, i\neq M$. 
Hence, we can write for $\epsilon_1, ..., \epsilon_k > 0$ as shown in equation(\ref{eq:supp},\ref{eq:supp1},\ref{eq:supp2}).

Now, for $\tau \rightarrow 0$ we have that $\alpha_M \rightarrow 1$ and for all other $i \neq M$ we have that $\alpha_i \rightarrow 0$. Hence, the probability distribution $\bm{\alpha}$ of the \srrn is always the one-hot encoding of a particular centroid.  

By an argument analog to the one we have made for Theorem \ref{theorem-pdfa}, we can prove that for every state $q\in\mathcal{Q}$ and every input token $a \in \Sigma$, the probability distribution $\bm{\alpha}$ of the \srrn is the same irrespective of the previous input sequences and visited states. Finally, by plugging in the one-hot encoding $\bm{\alpha}$ in both equations~\ref{eqn-transition-1} and \ref{eqn-transition-2}, we can conclude that the transition function of an \srrn without $\infty$-memory is identical to that of a DFA, because we always chose exactly one new state. 
\end{proof}

\end{document}